\pdfoutput=1
\documentclass[english]{article}
\usepackage[T1]{fontenc}
\usepackage[latin9]{inputenc}
\usepackage{geometry}
\usepackage{color}
\usepackage{array}
\usepackage{float}
\usepackage{rotfloat}
\usepackage{multirow}
\usepackage{amsmath}
\usepackage{amssymb}
\usepackage{graphicx}
\usepackage{setspace}
\usepackage[authoryear]{natbib}

\makeatletter

\providecommand{\tabularnewline}{\\}
\floatstyle{ruled}
\newfloat{algorithm}{tbp}{loa}
\providecommand{\algorithmname}{Algorithm}
\floatname{algorithm}{\protect\algorithmname}

\usepackage{amsfonts}
\usepackage{multicol}
\usepackage{threeparttable}
\usepackage{graphicx}
\usepackage{lscape}
\usepackage{placeins}

\usepackage{algorithm}
\usepackage[noend]{algpseudocode}

\@ifundefined{showcaptionsetup}{}{%
 \PassOptionsToPackage{caption=false}{subfig}}
\usepackage{subfig}
\makeatother

\usepackage{babel}
\begin{document}
\begin{spacing}{0}

\title{\noindent Causally Driven Incremental Multi Touch Attribution Using
a Recurrent Neural Network \thanks{The authors are part of \texttt{JD Intelligent Ads }Lab. The views
represent that of the authors, and not \texttt{JD.com}. Some aspects
of the data, institutional context and implementation are masked to
address business confidentiality. We thank Lixing Bo, Xi Chen, Jack
Lin and Paul Yan for their support, collegiality and collaboration
during the project.}}
\end{spacing}

\author{Ruihuan Du\thanks{\texttt{JD.com}}\and Yu Zhong$^{\dagger}$\and
Harikesh S. Nair\thanks{\texttt{JD.com} and Stanford University}\and
Bo Cui$^{\dagger}$\and Ryan Shou$^{\dagger}$}

\date{This draft: Jan 30, 2019}
\maketitle
\begin{abstract}
\begin{singlespace}
\noindent This paper describes a practical system for Multi Touch
Attribution (MTA) for use by a publisher of digital ads. We developed
this system for \texttt{JD.com}, an eCommerce company, which is also
a publisher of digital ads in China. The approach has two steps. The
first step (``response modeling'') fits a user-level model for purchase
of a product as a function of the user\textquoteright s exposure to
ads. The second (``credit allocation'') uses the fitted model to
allocate the incremental part of the observed purchase due to advertising,
to the ads the user is exposed to over the previous $T$ days. To
implement step one, we train a Recurrent Neural Network (RNN) on user-level
conversion and exposure data. The RNN has the advantage of flexibly
handling the sequential dependence in the data in a semi-parametric
way. The specific RNN formulation we implement captures the impact
of advertising intensity, timing, competition, and user-heterogeneity,
which are known to be relevant to ad-response. To implement step two,
we compute Shapley Values, which have the advantage of having axiomatic
foundations and satisfying fairness considerations. The specific formulation
of the Shapley Value we implement respects incrementality by allocating
the overall incremental improvement in conversion to the exposed ads,
while handling the sequence-dependence of exposures on the observed
outcomes. The system is under production at \texttt{JD.com}, and scales
to handle the high dimensionality of the problem on the platform (attribution
of the orders of about 300M users, for roughly 160K brands, across
200+ ad-types, served about 80B ad-impressions over a typical 15-day
period).
\end{singlespace}

\noindent \medskip{}

\begin{singlespace}
\end{singlespace}

\pagebreak
\end{abstract}
\baselineskip=1.5pc \parskip=0in

\pagebreak{}

\section{Introduction\label{sec:Introduction}}

As digital ads proliferate, and the measurability of advertising increases,
the issue of Multi Touch Attribution (MTA) has become one of paramount
importance to advertisers and digital publishers. MTA pertains to
the question of how much the marketing touchpoints a user was exposed
to, contributes to an observed action by the consumer. Understanding
the contribution of various marketing touchpoints is an input to good
campaign design, to optimal budget allocation and for understanding
the reasons for why one campaign worked and one did not. Wrong attribution
results in misallocation of resources, inefficient prioritization
of touchpoints, and consequently lower return on marketing investments.
Consequently having a good model of attribution is now recognized
as critical for marketing planning, design and growth. According to
eMarketer estimates, among US companies with at least 100 employees
using more than one digital marketing channel, about 85\% utilize
some form of digital attribution models in 2018, emphasizing the importance
of having good solutions to the problem from the perspective of industry
(\citealp{eMart2018}). 

Because the various touchpoints can interact in complex ways to affect
the final outcome, the problem of parsing the individual contributions
and allocating credit is a complex one. Given the complexity, many
firms and platforms use \textit{rule-based} methods such as last-touch,
first-touch, equally-weighted, or time-decayed attribution (\citealp{IAB2018}).
Because these rules may not always reflect actuality, modern approaches
propose \textit{data-driven} attribution schemes that use rules derived
from actual marketplace data to allocate credit. This paper proposes
a data-driven MTA system for use by a publisher of digital ads.\footnote{While the model is presented from the perspective of a publisher (in
our case, an eCommerce platform), one can also see this from the perspective
of an advertiser who wishes the assign credit for the orders he obtains
across the various ads he buys. } We developed this system for \texttt{JD.com}, an eCommerce company,
which is also a publisher of digital ads in China. The advertising
marketplace of JD feature thousands of advertisers buying billions
of impressions of ads of more than 200 types for over 300m users,
and is a data-rich environment. Hence, as a practical matter, we need
a system that scales to handle high dimensionality and leverages the
large quantities of user-level data available to the platform.

Our approach has two steps. The first step (``response modeling'')
fits a user-level model for purchase of a brand's product as a function
of the user\textquoteright s exposure to ads. The second (``credit
allocation'') uses the fitted model to allocate the incremental part
of the observed purchase due to advertising, to the ads the user is
exposed to over the previous $T$ days. 

To implement step one, our goal is to develop a response model that
captures the following aspects of ad-response, 
\begin{enumerate}
\item Responsiveness of current purchases \textit{to a sequence} of past
advertising exposures: that is, we would like to develop a response
model that allows for both current and past advertising to matter
in driving current purchases. This is consistent with a large literature
that has documented that advertising has long-lived effects, and emphasized
how both the stock as well as the flow of advertising affects consumer
response (e.g., \citealp{BAGWELL20071701}). In addition, various
findings in this literature motivates the need to handle the effect
of the history of past exposures flexibly. For instance, the effect
of history can operate in complex ways, by changing not just the baseline
level of purchase probability, but also the marginal effect of current
ad-exposures (e.g., \citealp{bassetal07,naiketal98}). 
\item Responsiveness of current purchases\textit{ to the intensity} of ad-exposure:
that is, the model should allow the effect of a brand's advertising
on a user to depend on the number of exposures, not just whether there
was exposure, with the effect being possibly non-linear (e.g., \citealt{DubeHitschManchanda05}).
Therefore, we need to capture an intensive margin by which advertising
can affect behavior, in addition to accommodating an extensive margin
for the effects of ads. 
\item Responsiveness of current purchases\textit{ to the timing} of ad-exposures:
that is, the model should allow the effect of past exposures to differ
based on the timing of those exposures. This is motivated by accumulating
evidence that the effect of advertising exposure is long-lasting but
decays in human memory (e.g., \citealp{Sahni2015}). Therefore, we
expect advertising to have a lasting effect, with the effect highest
at the time of exposure, \textit{ceteris paribus}, and decaying over
time. A realistic model should accommodate a role for time in incorporating
the effect of an exposure on purchase, and allow this decay to occur
in a flexible way that can be learned from the data.
\item Responsiveness of current purchases\textit{ to competitive} ad-exposures:
that is, the model should accommodate a role for both own and competing
brand's current and past ad-exposures to affect current purchases.
Allowing for cross-brand ad-effects to matter is important in a competitive
marketplace with competing brands, and is also critical to capturing
the incremental contribution to a brand's own-advertising efforts
correctly (e.g., \citealp{AndersonSimester2013}). 
\item Capturing \textit{heterogeneity across users}: that is, the model
should allow for the ad-response to differ by consumer types. One
motivation is based on business considerations: advertisers use the
output from the model to design their targeting strategies, and often
desire estimates of attribution split by consumer segments. Another
motivation is based on inference: a typical concern about measuring
ad-response is user selection into exposure. Including a flexible
accommodation for heterogeneity into the model mitigates the selection
concern somewhat by ``controlling for'' observables that drive selection
into exposure (e.g., \citealp{Varian7310}).
\end{enumerate}
Given the scale of the data, and the large number of ad-types, a fully
non-parametric model that reflect these considerations is not feasible.
Instead, our approach is to develop a flexible specification that
fits the data and incorporates these aspects of ad-response. We train
a Recurrent Neural Network (RNN) for this purpose. The RNN is trained
on user-level conversion and exposure data, and is architected to
capture the impact of advertising intensity, timing, competition,
and user-heterogeneity outlined above. The model is set up as a binary
classification problem, outputting a probability that a user buys
a product associated with a brand in a given time-period. It takes
as inputs in its lowermost layer, the impressions served to a user
of a focal brand's and its competitors ads split by ad-type; and separately
for each of $T$ time-periods prior to date of attribution. This allows
a flexible way of handling the intensity of ad-exposure and competition
over the past $T$ periods on attribution of an order on the $T^{\textrm{th}}$
period. A separate fully connected layer in the model takes as input
a set of user characteristics, which shifts the the ``intercept''
of the logistic output layer, giving it a ``fixed-effects'' interpretation. 

The specifics of the application to advertising-response motivates
a bi-directional formulation of the RNN. We allow for a hidden layer
with backward recurrence, augmented with a hidden layer with forward
recurrence. This improves the fit of the model, because the fact that
a user $i$ saw a particular set of ads in $t+1,..,T$ is useful to
predict his response in period $t$, $Y_{it}$. For example, if a
user bought a brand in $t$, he may not search for that brand in period
$t+1$, and not be exposed to search ads in $t+1$. So the knowledge
that he did not see search ads in $t+1$ is useful to predict whether
he will buy in period $t$. More generally, this suggests that the
sequence of future ad-impressions $\mathbf{x}_{i,t+1:T}$ can help
predict $Y_{it}$. Adding a layer with forward recurrence serves as
a semi-parametric summary of future activity that is helpful to predict
current actions. 

While RNNs are not new to this paper, it is worth emphasizing why
this class of models is of value for the MTA problem. Compared to
the other frameworks, RNNs represent a more flexible way to handle
the sequential dependence in the data. Sequential dependence is key
to ad-response, because what we need to capture from the data is how
exposure to past touchpoints cumulatively build up to affect the final
outcome. RNNs do this well by allowing for continuous, high-dimensional
hidden states (compared to lower-dimensional, discrete ones in other
models with hidden states); combined with a distributed representation
of those states that allows them to store information about the past
efficiently. This enables the RNN to handle long-term, higher-order
and non-Markovian dependencies in a semi-parametric manner (see \citealt{Graves2012,Lipton15}
for overview).\footnote{Comparing Hidden Markov Models to RNNs, \citep{Lipton15} says, ``Hidden
Markov models (HMMs), which model an observed sequence as probabilistically
dependent upon a sequence of unobserved states, were described in
the 1950s and have been widely studied since the 1960s \citep{Stratonovich60}.
However, traditional Markov model approaches are limited because their
states must be drawn from a modestly sized discrete state space $S$.
The dynamic programming algorithm that is used to perform efficient
inference with hidden Markov models scales in time $O(\left|S\right|^{2})$
\citep{Viterbi50}. Further, the transition table capturing the probability
of moving between any two time-adjacent states is of size $\left|S\right|^{2}$.
Thus, standard operations become infeasible with an HMM when the set
of possible hidden states grows large. Further, each hidden state
can depend only on the immediately previous state. While it is possible
to extend a Markov model to account for a larger context window by
creating a new state space equal to the cross product of the possible
states at each time in the window, this procedure grows the state
space exponentially with the size of the window, rendering Markov
models computationally impractical for modeling long-range dependencies
\citep{Gravesetal14}.'' Increasingly, some researchers view HMMs
as special cases of RNNs (\citep{WesselsOmlin2000,Buys2018}).} Well-known results in theoretical computer science also show that
recurrent neural nets have attractive universal approximation properties:
any function that can be computed by a digital computer is also in-principle
computable by a recurrent neural net architecture.\footnote{By a ``net'' we mean, an architecture in which neurons are allowed
to synchronously update their states according to some combination
of past activation values. While earlier literature had suggested
that nets can achieve universality if one allowed for an infinite
number of neurons (e.g., \citet{FrabklinGarzon90}), or allowed for
higher-order connections (where current states update their values
as multiplications or products of past activations, e.g., \citet{sunetal91}),
results by \citet{SiegelmannSontag91,SiegelmannSontag95} are even
more favorable: recurrent neural nets can achieve universality using
only a finite number of neurons, and using only first-order, non-multiplicative
connections. In particular, any function computable by a Turing Machine
can be computed by such a net. ``Turing Machines'' are mathematically
simple computational devices to help formalize the notion of computability.
Under the \textit{\footnotesize{}Church-Turing} thesis in computer
science, for every computable problem, there exists a Turing Machine
that computes it; and conversely, every problem not computable by
a Turing Machine is also not computable by finite means. See for instance
\texttt{\footnotesize{}https://plato.stanford.edu/entries/turing-machine/}{\footnotesize{},
for historical perspectives.}} Thus, relatively ``simple'' recurrent architectures can capture
very complex functional dependencies in the data, especially if we
allow for a large number of neurons. This makes them attractive to
our situation. The main disadvantage of RNNs are they require more
data, and take longer to train. This problem is mitigated in implementations
in modern tech platforms, which are data-rich and have access to large
computational resources.

To implement step two, we focus on incrementality-based allocation
for advertising. To understand the motivation for this, note that
each ad generates an incremental increase in the overall probability
of purchase, and the set of ad-exposures as a whole generate an incremental
improvement in the propensity to purchase. Our approach is to allocate
the incremental improvement due to the ads to each ad-type. This takes
into account that even if the user did not see the ads, the user may
have some baseline propensity to buy anyway due to tastes, prior experiences,
spillovers from competitive advertising. Logically, the part of observed
orders that would have occurred anyway should not be allocated to
the focal brand's advertising efforts. To allocate credit, we compute
Shapley Values, which have the advantage of having axiomatic foundations
and satisfying fairness considerations (\citealp{Shapley53,roth_1988}).
The specific formulation of the Shapley Value we implement respects
incrementality by allocating the overall incremental improvement in
conversion to the exposed ads, while handling the sequence-dependence
of exposures on the observed outcomes. 

Computing the Shapley Values is computationally intensive. We present
a scalable algorithm (implemented in a distributed \texttt{MapReduce}
framework) that is fast enough to allow computation in reasonable
amounts of time so as to make productization feasible. The algorithm
takes predictions from the response model trained on the data as an
input, and allocates credit over tuples of ad-exposures and time periods.
Allocation at the tuple-level has the advantage of handling the role
of the sequence in an internally consistent way. Once allocation of
credit at this level is complete, we sum across all time periods associated
with an ad-type to develop an \textit{ex-post} credit allocation to
each ad-type. This explicit aggregation has the advantage that aggregation
biases are reduced when using the model to allocate credit at a more
aggregate level, such as over advertising channels (e.g., search and
display).\footnote{Aggregating responses to the channel level reduces the complexity
of the algorithm, and enables pooling of data, but masks the differential
contribution of various touchpoints to final conversion, because implicitly,
such a response model assumes that the effect of all touchpoints within
the channel are similar. By training the response model and implementing
the allocation at the ad-type and time-period level, and then exactly
aggregating these allocations to the channel level, we mitigate such
concerns to a large extent.} 

In combination, the RNN response-model and Shapley Value credit system
represents a coherent, theory- and data-driven attribution framework
for the platform. We present details and an illustration of the framework
using data from one product category (cell-phones) at \texttt{JD.com}.
This is a single product category version of the full framework. The
full framework is under production at the firm, and accommodates all
product categories on the site, and scales to handle the high dimensionality
of the problem on the platform (attribution of the orders of about
300M users, for roughly 160K brands, across 200+ ad-types, served
about 80B ad-impressions over a typical 15-day period).

The rest of the paper discusses the relevant literature, the details
of the model, details of the cell-phone data, and results. The last
section concludes.

\section{Relationship to the Literature}

The problem of attribution of credit to underlying marketing activities
is not new. The previous literature on the topic is divided into two
streams: (a) empirical papers that develop statistical response models
to measure the effect of marketing touchpoints on consumer purchase
behavior and engagement; (b) papers that combine an empirically specified
response model with an allocation scheme for allocating credit to
the touchpoints. This paper is part of the second stream.

Early research on response models used \textit{market-level data}
and time-series based \textit{aggregate ``marketing mix'' models}
to asses the effect of marketing touchpoints in print, TV  and internet
channels on sales and engagement (e.g., \citealp{Nairketal2005,DEHAAN2016491,KIREYEV2016475}).
More recent work has leveraged access to \textit{user-level} browsing
and conversion data to develop \textit{individual-level models} of
responsiveness. Notable examples in this stream include \citet{Shao2011}
(who use a bagged logistic regression model and a semi-parametric
model that allows upto second-order dependence in consumer behavior);
\citet{LiKannan2014} and \citet{XuDuanWhinston14} (who use customized
Markovian models of consumer channel choice and conversion); \citet{ZhangWeiRen2014}
(who use a hazard-based survival model that allows for time decay
in ad-exposures); \citet{Abhishek15} (who use an HMM of user exposure
and conversion); and \citet{ANDERL2016457} who model customer purchase
and browsing behavior as Markov graph with upto fourth order dependence. 

Broadly speaking, the recent response modeling literature has focused
on developing generative models of consumer behavior that capture
the dependence in the effects of touchpoints to the extent possible,
while making simplifying assumptions to feasibly handle the high dimensionality
of the measurement problem. Relative to this stream, this paper uses
an RNN trained on user-level data as a response model. Compared to
past frameworks, the model handles complex patterns of dependence
in a more flexible way. The specific formulation of the RNN also allows
the sequences of touchpoints to have differential effects on final
conversion, which is novel. The setup also accommodates in one framework
the role of intensity, timing, competition, and user-heterogeneity,
which have typically not been considered together in the previous
literature. 

Amongst papers in the second stream, \citet{Dalessandro2012} was
the first to propose using the Shapley value as a credit allocation
mechanism for the MTA problem. They call this ``causally-motivated''
attribution because of the causal interpretation associated with the
``marginality'' property of the Shapley Value rule. \citet{Dalessandro2012}
compute Shapley values for channels by fitting to the data logistic
regression models similar to \citet{Shao2011}. \citet{Yadagiri15}
present an important extension this work, allowing the statistical
model to be semi-parametric, but restricting outcomes to depend on
the composition, but not the order, of previous touchpoints. \citet{ANDERL2016457}
leverages the Markov graph-based approach proposed by \citet{ArchakMirrokni2010Muthukrishnan}
for credit allocation. This approach is computationally attractive,
but lacks the fairness properties of the Shapley Value.\footnote{The credit allocated to an ad-slot $s$ is the ``removal effect'',
computed as the change in probability in reaching the \texttt{conversion}
state from the \texttt{start} state when the slot $s$ is removed
from the Markov graph. The removal effect can be thought of as the
marginal contribution of an ad-slot. The Shapley Value in contrast,
allocates credit based on a transformation of the marginal contributions.} Like \citet{Shao2011,Dalessandro2012,Yadagiri15}, we use the Shapley
value for credit allocation. The specifics of our implementation differs
from these papers on three aspects. First, we present a way to obtain
the incremental contribution from a focal firms' advertising to observed
orders, and to allocate the incremental contribution to the underlying
ad-slots. Previous approaches have allocated total orders. Our view
is that the incremental part is the more intuitive allocation as it
is the component that is due to advertising. Second, allowing the
conversion to depend on the order of exposures in the response model
requires us to develop a way to implement credit allocation to an
ad-slot which depends on its order in the temporal sequence of exposures.
We present an algorithm that computes Shapley Values over tuples of
ad-slots and location in the sequence to do so. This aspect, which
arises because ``order matters'', is not an explicit consideration
in previous approaches. Third, we implement attribution at a more
disaggregated ``ad-slot'' level, compared to the more aggregate
channel-level attribution of past approaches. This makes the problem
considered here more high dimensional than considered previously.

This paper is also related to a game-theoretical literature that devise
payment rules for multi-channel ads. Notable papers include \citet{AgarwalAtheyYang09,WiburZhu09,Jordan2011,HuShinTang2016,Berman2018}
who propose efficient contracts when there are interactions across
publishers, and advertisers and publishers are strategic and information
is possibly asymmetric. The response model and attribution methods
presented here can form an input to the creation of the payment contracts
suggested in this theory.

A limitation of our approach and indeed of all the response models
cited previously, is the lack of exogenous variation in user exposure
to advertising. This can contaminate the learning of marginal effects
from the data due to issues associated with nonrandom targeting and
selection into ad-exposure. A typical solution to the problem $-$
randomization of users into ad-exposures across all the ad-types,
followed by training the model on data generated by the randomization$-$
is infeasible at the scale required for practical implementation,
due to the cost and complexity of such randomization. Extant papers
that have trained ad-response models on data with full or quasi-randomization
(\citealp{Sahni2015,BarajasAkella2016,Nairetal2017,ZantedeschiBradlow2017})
have done so at smaller scale, over limited number of users and ad-types.
The approach to this problem adopted here is to include a large set
of user features into the model so that by including these, we convert
a ``selection on unobservables'' problem into a ``selection on
observables'' problem. Given the feature set is large and accommodated
flexibly, controlling for these observables may mitigate the selection
issue to a great extent (e.g., \citealp{Varian7310}), albeit not
perfectly.

\section{Model Framework\label{sec:Model-Framework}}

\subsection{Problem Statement: Defining Multi Touch Attribution in terms of Incrementality}

Let $i=1,..,N$ denote users; $t=1,..,T$ denote time (days); and
$b=1,..,B$ denote brands. Let $k=1,..,K$ index an ``\textit{ad-position},''
i.e., a particular location on the publishers inventory or at an external
site at which the user can see advertisements linked to a given brand.
For instance, a particular ad-slot showing a display ad on the top
frame of the \texttt{JD} app home-page would be one ad-position, and
a particular ad-slot showing a search-ad in in response to keyword
search on the \texttt{JD} app would be another ad-position. Consider
an order $o\left(i,b,T\right)$ made by user $i$ for brand $b$ on
day $T$. Let $\mathcal{A}_{ibT}\subseteq K$ denote the set of ad-positions
at which user $i$ was exposed to ads for brand $b$ over the $T$
days preceding the order (from $t=1$ to $T$). 

We formulate the multi touch attribution problem as developing a set
of credit-allocations $\varrho_{k}\text{\ensuremath{\left(\mathcal{A}_{ibT}\right)}}$
for all $k\in\mathcal{A}_{ibT}$, so that the allocation for $k$
represents the contribution of brand $b$'s ads at position $k$ to
the expected \textit{incremental }benefit generated by brand $b$'s
advertising on the observed order. Define $v\left(\mathcal{A}_{ibT}\right)$
as the change in the probability of order $o\left(i,b,T\right)$ occurring
due to the user's exposure to \textbf{$b$}'s ads in the positions
in $\mathcal{A}_{ibT}$. We look for a set of fractions $\varrho_{k}\text{\ensuremath{\left(\mathcal{A}_{ibT}\right)}}$
such that,
\begin{equation}
\sum_{k\in\mathcal{A}_{ibT}}\varrho_{k}\text{\ensuremath{\left(\mathcal{A}_{ibT}\right)}}=v\left(\mathcal{A}_{ibT}\right).\label{eq:allocative-efficiency}
\end{equation}

\subsection{Problem Solution: Response Model Trained on User Data + Shapley Values}

\begin{figure}[p]
\begin{centering}
\centering \caption{Attribution System Architecture\label{fig:AttributionSystemArchitecture}}
\par\end{centering}
\begin{centering}
\includegraphics[scale=0.45]{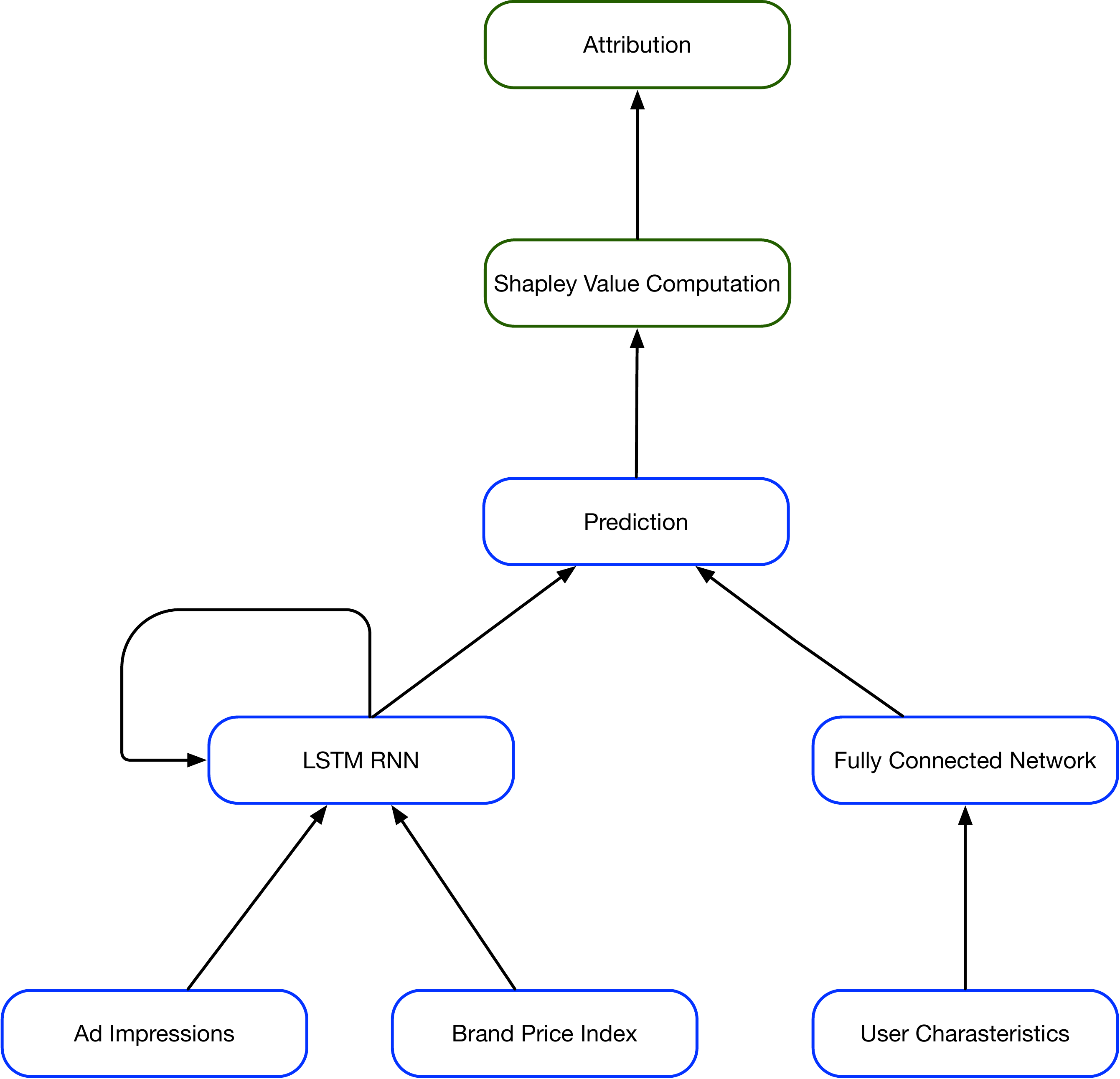}
\par\end{centering}
\begin{spacing}{0.9}
\centering{}\begin{threeparttable}\begin{tablenotes} \item \textit{\scriptsize{}Notes:}{\scriptsize{}
The Figure shows the architecture of the attribution system. In the
first stage, a response-model is trained on historical data. In the
response model, ad-impressions and price indices are used as inputs
into an LSTM layer with recurrence. User characteristics are processed
through a fully connected layer. These together feed into an output
layer that forms a prediction. Results from the trained model are
used to compute Shapley values for all orders observed at a disaggregated
level. These are then aggregated as desired to obtain attributions.}\end{tablenotes}
\end{threeparttable} 
\end{spacing}
\end{figure}

We solve the problem in two steps. To allocate the orders on date
$T$, 
\begin{itemize}
\item In step 1, we train a response model for purchase behavior using individual
user-level data observed during $t=1$ to $T$.
\item In step 2, we take the model as given, and for each order $o\left(i,b,T\right)$
observed on date $T,$ we compute Shapley Values for the ad-positions
$k\in\mathcal{A}_{ibT}$. We set $\varrho_{k}\text{\ensuremath{\left(\mathcal{A}_{ibT}\right)}}$
to these Shapley values and aggregate across orders to obtain the
overall allocation for brand $b$ on date $T$.
\end{itemize}
Figure (\ref{fig:AttributionSystemArchitecture}) show the architecture
of the system.

\subsubsection{Shapley Values}

\paragraph{Motivation}

The Shapley value has many advantages as a fair allocation scheme
in situations of joint generation of outcomes. These advantages have
been articulated in a long history of economic literature on co-operative
games starting with \citet{Shapley53}. See \citet{roth_1988} for
a summary and historical perspectives. The key idea is that any fair
allocation should avoid waste and allocate all of the total benefit
to the constituent units that generated the benefit (``allocative
efficiency''). This is the ``adding-up'' requirement is encapsulated
in equation (\ref{eq:allocative-efficiency}). Further, fairness suggests
that two units that contribute the same to every possible configuration
in which the constituent units can influence the final outcome, should
be allocated the same credit (``symmetry of credit''); and logically,
that a unit which contributes nothing to any configuration should
be allocated zero credit (``dummy unit''). 

In addition to these requirements, a fair allocation should also satisfy
the ``marginality principle,'' encapsulating the idea that credit
should be proportional to contribution. Specifically, marginality
requires that the share of total joint benefit allocated to any constituent
unit should depend only on that unit's own contribution to the joint
benefit. As \citet{young_1988} points out, a sharing rule that does
not satisfy the marginality principle is subject to serious distortions.
If one unit's credit depends on another\textquoteright s contributions,
the first unit can be viewed favorable (or unfavorably) on the basis
of the performance of the second. This affects how these units are
rewarded or punished and distorts performance. The difficulty is that
finding a sharing rule that simultaneously satisfies marginality and
efficiency is non-trivial. When the marginality principle is imposed,
the sum of individual unit\textquoteright s marginal contributions
will not equal total overall benefit in general. If there are increasing
returns from joint production, the sum of marginal contributions will
be too high; and if there are decreasing returns, it will be to low.\footnote{Paraphrasing \citet{young_1988}: ``One seemingly innocuous remedy
is to compute the marginal product of all factor inputs and then adjust
them by some common proportion so that total output is fully distributed.
This proportional to marginal product principle is the basis of several
classical allocation schemes..{[}{]}..the proportional to marginal
product principle does not resolve the ``adding up'' problem in
a satisfactory way. The reason is that the rule does not base the
share of a factor solely on that factor's own contribution to output,
but on all factors contribution to output. For example, if one factor's
marginal contribution to output increases while another \textquoteright s
decreases, the share attributed to the first factor may actually decrease;
that is, it may bear some of the decrease in productivity associated
with the second factor.''} 

The Shapley value aligns these requirements in an elegant way. Theorem
1 in \citet{young_1988} shows the remarkable property that the Shapley
value is the \textit{unique} sharing rule that is efficient, symmetric
and satisfies the marginality principle. This is the reason for the
Shapley value's appeal as a credit-allocation scheme.

The next three sub-sections discusses how we compute the Shapley values
in the ad-position attribution problem. First, we discuss how we define
the expected incremental benefit generated by a set of ad-positions
to a brand's order. Then, we discuss how we allocate that benefit
to ad-position-day tuples, and aggregate these to generate credit
allocations across ad-positions. Finally, we provide an illustrative
example.

\paragraph*{Defining the Expected Incremental Benefit Generated by a Set of Ad-Positions
to an Observed Order}

Let $Y_{ibT}\in\left(0,1\right)$ denote a binary random variable
for whether user $i$ purchases brand $b$ on day $T$. An order $o\left(i,b,T\right)$
is a realization $Y_{ibT}=1$ with associated own-brand ad-exposures
at positions $\mathcal{A}_{ibT}$. The expected\textit{ }incremental
benefit generated by brand $b$'s advertising on order $o\left(i,b,T\right)$
is,
\begin{alignat}{1}
v\left(\mathcal{A}_{ibT}\right) & =\mathbb{E}\left[Y_{ibT}|\mathcal{A}_{ibT}\right]-\mathbb{E}\left[Y_{ibT}|\varnothing_{b}\right]\label{eq:Total-incremental-benefit}
\end{alignat}

The first term in equation (\ref{eq:Total-incremental-benefit}) represents
the probability of the order $o\left(i,b,T\right)$ occurring given
$i$'s exposures to brand $b$'s ads at the ad-positions in $\mathcal{A}_{ibT}$
over the $T$ preceding days. The second term represents the counterfactual
probability of the order occurring if $i$ \textit{did not have} any
exposures to brand $b$'s ads at the ad-positions in $\mathcal{A}_{ibT}$
over the $T$ preceding days (denoted as $\varnothing_{b}$). Holding
everything else fixed, this difference represents the expected incremental
contribution of the ad-positions in $\mathcal{A}_{ibT}$ to the order.
We can think of $v\left(\mathcal{A}_{ibT}\right)$ as a causal effect
of brand $b$'s advertising over the past $T$ days on user $i'$s
propensity to place the observed order on day $T.$ 

\paragraph*{Allocating Incremental Benefit to a Position-Day Tuple\label{par:Allocating-Incremental-Benefit}}

To allocate $v\left(\mathcal{A}_{ibT}\right)$ to the ad-positions
in $\mathcal{A}_{ibT}$, we first allocate $v\left(\mathcal{A}_{ibT}\right)$
to each \textit{ad-position-day tuple} in which $i$ saw ads of brand
$b$ over the last $T$ days. We then sum the allocations across days
for the tuples that each ad-position $k\in\mathcal{A}_{ibT}$ is associated
with, to obtain the overall allocation of $v\left(\mathcal{A}_{ibT}\right)$
to that $k$.

To do this, let $\mathcal{N}_{ibT}$ be the set of ad-position-day
combinations at which user $i$ saw ads for brand $b$ during the
$T$ days preceding order $o\left(i,b,T\right)$. Denote the cardinality
of $\mathcal{N}_{ibT}$ as $\left|\mathcal{N}_{ibT}\right|$.\footnote{For example, if user $i$ saw ads for brand $b$ at ad-position $k=5$
on days $t=2,3$; at position $k=18$ on day $t=10$; and position
$k=102$ on day $t=15$, $\mathcal{N}_{ibT}$ would be $\left(\left\{ 5,2\right\} ,\left\{ 5,3\right\} ,\left\{ 18,10\right\} ,\left\{ 102,15\right\} \right)$
and $\left|\mathcal{N}_{ibT}\right|=4$.} For a given tuple $\left\{ k,t\right\} $, let $S$ denote a generic
element from the power-set of $\mathcal{N}_{ibT}\left\backslash \left\{ k,t\right\} \right.$,
i.e., the sub-set of all ad-position-day combinations at which the
user saw ads for brand $b$ during the $T$ days, excluding tuple
$\left\{ k,t\right\} $. Let the cardinality of $S$ be denoted $\left|S\right|$. 

Define the function $w\left(S\right)$ as,
\begin{equation}
w\left(S\right)=\mathbb{E}\left[Y_{ibT}|S\right]-\mathbb{E}\left[Y_{ibT}|\varnothing_{b}\right]\label{eq:incremental-benefit-for-S}
\end{equation}
i.e., $w\left(S\right)$ represents the expected incremental benefit
from user $i$ seeing ads for brand $b$ at ad-position-day combinations
in $S$, holding everything else fixed. By construction, $w\left(\mathcal{N}_{ibT}\right)=v\left(\mathcal{A}_{ibT}\right)$.
So, by allocating $w\left(\mathcal{N}_{ibT}\right)$ across the ad-position-day
tuples in $\mathcal{N}_{ibT}$, we allocate the same total incremental
benefit generated by brand $b$'s advertising, as we would by allocating
$v\left(\mathcal{A}_{ibT}\right)$ across the ad-positions in $\mathcal{A}_{ibT}$.
Also, by construction, $w\left(\varnothing_{b}\right)=0$.

For each tuple $\left\{ k,t\right\} \subseteq\mathcal{N}_{ibT}$,
we need the fractions $\varrho_{\left\{ k,t\right\} }\text{\ensuremath{\left(\mathcal{N}_{ibT}\right)}}$,
to satisfy two conditions. First, that $\sum_{\left\{ k,t\right\} \subseteq\mathcal{N}_{ibT}}\varrho_{\left\{ k,t\right\} }\text{\ensuremath{\left(\mathcal{N}_{ibT}\right)}}=w\left(\mathcal{N}_{ibT}\right)$,
so that the fractions sum to the full incremental benefit of the ads
on the order (i.e., satisfy allocative efficiency). Second, that the
fractions for a given tuple $\left\{ k,t\right\} $ are functions
of only its marginal effects with respect to $w\left(.\right)$ (i.e.,
satisfy the marginality principle). These are the Shapley values for
the tuples $\left\{ k,t\right\} \subseteq\mathcal{N}_{ibT}$ defined
as,

\begin{equation}
\varrho_{\left\{ k,t\right\} }\text{\ensuremath{\left(\mathcal{N}_{ibT}\right)}}=\sum_{S\subseteq\mathcal{N}_{ibT}\left\backslash \left\{ k,t\right\} \right.}\frac{\left|S\right|!\left(\left|\mathcal{N}_{ibT}\right|-\left|S\right|-1\right)!}{\left|\mathcal{N}_{ibT}\right|!}\left[w\left(S\cup\left\{ k,t\right\} \right)-w\left(S\right)\right]\label{eq:Shapley-value-k-t}
\end{equation}
Computing the Shapley values requires a way to estimate the marginal
effects $w\left(S\cup\left\{ k,t\right\} \right)-w\left(S\right)$
in equation (\ref{eq:Shapley-value-k-t}) from the data, as well as
an algorithm that scales to handle the high-dimensionality of $S$.
This is discussed in the subsequent section. 

Once the Shapley values $\varrho_{\left\{ k,t\right\} }\text{\ensuremath{\left(\mathcal{N}_{ibT}\right)}}$
are computed, we sum them across all $t$ to obtain the allocation
of that order to ad-position $k$ as,
\begin{equation}
\varrho_{k}\text{\ensuremath{\left(\mathcal{N}_{ibT}\right)}}=\sum_{t\subseteq\mathcal{N}_{ibT}\left(k\right)}\varrho_{\left\{ k,t\right\} }\text{\ensuremath{\left(\mathcal{N}_{ibT}\right)}}\label{eq:Shapley-value-k}
\end{equation}
where $\mathcal{N}_{ibT}\left(k\right)$ is the set of days in $\mathcal{N}_{ibT}$
that are associated with ad-position $k.$ 

The final step is to do this across all orders observed for brand
$b$ on day $T$. To do this, we sum $\varrho_{k}\text{\ensuremath{\left(\mathcal{N}_{ibT}\right)}}$
across all $k$ and all users who bought brand $b$ on day $T$. This
gives the overall incremental contribution of the $K$ ad-positions
to the brand's orders. To allocate this to $k,$ we simply compute
how much ad-position $k$ contributed to this sum. To see this mathematically,
denote $\varrho_{k}\text{\ensuremath{\left(\mathcal{N}_{ibT}\right)}}$
in equation (\ref{eq:Shapley-value-k}) as $\varrho_{ibkT}$ for short.
We sum $\varrho_{ibkT}$ across all $k$ for all $i$ that made an
order (i.e., $Y_{ibT}=1$) to compute the term in the denominator
in equation (\ref{eq:Overall-proortional-allocation}); and we sum
$\varrho_{ibkT}$ for only ad-position $k$ across all $i$ that made
an order (i.e., $Y_{ibT}=1$) to compute the term in the numerator
in equation (\ref{eq:Overall-proortional-allocation}). Dividing the
numerator by the denominator, we allocate to ad-position $k$, a proportion
$\varPsi_{bkT}$ computed as, 
\begin{equation}
\varPsi_{bkT}=\frac{\sum_{\left\{ i:Y_{ibT}=1\right\} }\varrho_{ibkT}}{\sum_{k}\sum_{\left\{ i:Y_{ibT}=1\right\} }\varrho_{ibkT}}\label{eq:Overall-proortional-allocation}
\end{equation}
Each element $k$ in $\mathbf{\boldsymbol{\varPsi}}_{bT}=\left(\varPsi_{b1T},..,\varPsi_{bKT}\right)$
represents the contribution of ad-position $k$ to the total incremental
orders obtained on day $T$ by the brand due to its advertising on
the $K$ positions. $\mathbf{\boldsymbol{\varPsi}}_{bT}$ thus represent
a set of attributions that can be reported back to the advertiser.

\paragraph*{Linking to a Response-Model}

Let $x_{ibkt}$ be the number of impressions of brand $b$'s ad seen
by user $i$ at ad-position $k$ on day $t.$ Collect all the impressions
of the user for the brand's ad across positions on day $t$ in $\mathbf{x}_{ibt}=\left(x_{ib1t},..,x_{ibKt}\right)$;
collect the impression vectors across all the brands for that user
on day $t$ in $\mathbf{x}_{it}=\left(\mathbf{x}_{i1t},..,\mathbf{x}_{iBt}\right)$;
and stack the entire vector of impressions across all days and brands
in a $\left(K\cdot B\cdot T\right)\times1$ vector $\mathbf{x}_{i,1:T}=$$\left(\mathbf{x}_{i1},..,\mathbf{x}_{iT}\right)'$.
Let $p_{bt}$ be a price-index for brand $b$ on day $t$, representing
an average price for products of brand $b$ faced by users on day
$t$.\footnote{We compute this as a share weighted average of the list prices of
the SKUs associated with the brand on that day.} Collect the price indices for all brand on day $t$ in vector $\mathbf{p}_{t}=\left(\mathbf{p}_{1t},..,\mathbf{p}_{Bt}\right)$
and stack these in a $\left(B\cdot T\right)\times1$ vector $\mathbf{p}_{1:T}=\left(\mathbf{p}_{1},..,\mathbf{p}_{T}\right)'$.
Finally, let $\mathbf{d}_{i}$ represent a $R\times1$ vector of user
characteristics collected at baseline. The probability of purchase
on day $T$ is modeled as a function of user characteristics, and
the ad-impressions and price-indices of brand $b$ and all other brands
in the product category over the last $T$ days as,
\begin{equation}
\mathbb{E}\left[Y_{ibT}\right]=\Pr\left(Y_{ibT}=1\right)=\sigma\left(\mathbf{x}_{i,1:T},\mathbf{p}_{1:T},\mathbf{d}_{i};\hat{\varOmega}\right)\label{eq:Response-model}
\end{equation}
The probability model $\sigma\left(.\right)$ is parametrized by vector
$\hat{\varOmega}$ which will be learned from the data.\footnote{The ``hat'' notation on $\hat{\varOmega}$ emphasizes that the response
parameters $\hat{\varOmega}$ are learned in a first-stage from the
data.}

We use equation (\ref{eq:Response-model}) which provides an expression
for $\mathbb{E}\left[Y_{ibT}\right]$, along with the definition of
the marginal effects $w\left(S\right)$ in equation (\ref{eq:incremental-benefit-for-S}),
to compute the Shapley values defined in equation (\ref{eq:Shapley-value-k-t}).
To obtain the marginal effects from the response model, we define
an operator $\varGamma_{b}\left(.\right)$ on $\mathbf{x}_{i,1:T}$
that takes a set $S$ as defined in $\mathsection$\ref{par:Allocating-Incremental-Benefit}
as an input.\footnote{Recall from $\mathsection$\ref{par:Allocating-Incremental-Benefit}
that we use $S$ to refer to a sub-set of ad-position-day combinations
at which the user saw ads for brand $b$ during the $T$ days, excluding
tuple $\left\{ k,t\right\} $.} Given $S$, $\varGamma_{b}\left(.\right)$ sets all the impressions
of brand $b$ apart from those in the ad-position-day tuples in $S$
to $0.$ $\varGamma_{b}\left(.\right)$ also leaves impressions of
all brands $b'\neq b$ unchanged. 

Mathematically, taking $\mathbf{x}_{i,1:T}$ and $S$ as input, $\varGamma_{b}\left(\mathbf{x}_{i,1:T},S\right)$
outputs a transformed vector $\mathbf{x}_{i,1:T}^{\left(b,S\right)}$
computed as, 
\begin{equation}
x_{ib'kt}^{\left(b,S\right)}=\left\{ \begin{array}{ccc}
0 & \textrm{if} & b'=b\textrm{ and}\left\{ k,t\right\} \subsetneq S\\
x_{ib'kt} & \textrm{otherwise}
\end{array}\right.\label{eq:transformed-X}
\end{equation}
With $x_{ib'kt}^{\left(b,S\right)}$ as defined above, we can compute
the Shapley value using the response model as,
\begin{equation}
\varrho_{\left\{ k,t\right\} }\text{\ensuremath{\left(\mathcal{N}_{ibT}\right)}}=\sum_{S\subseteq\mathcal{N}_{ibT}\left\backslash \left\{ k,t\right\} \right.}\frac{\left|S\right|!\left(\left|\mathcal{N}_{ibT}\right|-\left|S\right|-1\right)!}{\left|\mathcal{N}_{ibT}\right|!}\left[\sigma\left(\mathbf{x}_{i,1:T}^{\left(b,S\cup\left\{ k,t\right\} \right)},\mathbf{p}_{1:T},\mathbf{d}_{i};\hat{\varOmega}\right)-\sigma\left(\mathbf{x}_{i,1:T}^{\left(b,S\right)},\mathbf{p}_{1:T},\mathbf{d}_{i};\hat{\varOmega}\right)\right]\label{eq:Shapley-vlaue-Using-response-model}
\end{equation}
In effect, what we obtain in the square brackets in equation (\ref{eq:Shapley-vlaue-Using-response-model})
is the change in the predicted probability of purchase of an order
of brand $b$ on day $T$ by user $i$ when the tuple $\left\{ k,t\right\} $
is added to the set of ad-position-day combinations in $S$, holding
everything else (including competitor advertising) fixed at the values
observed in the data for that order.

\paragraph*{Illustrative Example}

Suppose there are only two brands $\left(B=2\right)$, three ad-positions
$\left(K=3\right)$, and three days $\left(T=3\right)$. Suppose user
$i$ who made an order for brand $1$ on day $T=3,$ saw 4 ads for
brand $1$ at ad-position $k=1$ on days $t=2,3$; 7 ads at position
$k=2$ on day $t=2$; and 10 ads for brand 2 at ad-position 3 on day
$t=2$. Then, $\mathbf{x}_{i11}=\left(0,0,0\right)$, $\mathbf{x}_{i12}=\left(4,7,0\right)$,
$\mathbf{x}_{i13}=\left(4,0,0\right)$, $\mathbf{x}_{i21}=\left(0,0,0\right)$,
$\mathbf{x}_{i22}=\left(0,0,10\right)$, $\mathbf{x}_{i23}=\left(0,0,0\right)$,
so that $\mathbf{x}_{i,1:T}=$ (0,0,0,0,0,0,4,7,0,0,0,10,4,0,0,0,0,0$)'$.
Suppose, we would like to evaluate the Shapley value of tuple $\left\{ k,t\right\} =\left\{ 1,3\right\} $
of brand 1. For this order, $\mathcal{N}_{i1T}$ is the set $\left(\left\{ 1,2\right\} ,\left\{ 2,2\right\} ,\left\{ 1,3\right\} \right)$,
i.e., the $\left\{ k,t\right\} $ tuples at which the user saw ads
for brand $b=1$. The cardinality of the set $\left|\mathcal{N}_{ibT}\right|=3$.
$\mathcal{N}_{ibT}\left\backslash \left\{ 1,3\right\} \right.$ is
the set $\left(\left\{ 1,2\right\} ,\left\{ 2,2\right\} \right)$
with corresponding power set $\left(\left\{ \varnothing\right\} ,\left\{ 1,2\right\} ,\left\{ 2,2\right\} ,\left\{ \left\{ 1,2\right\} ,\left\{ 2,2\right\} \right\} \right)$.
In equation (\ref{eq:Shapley-vlaue-Using-response-model}) for the
Shapley values, $S$ is an element from this power set. To evaluate
the terms in the square brackets, we need to evaluate the probability
$\sigma\left(.\right)$ at transformed values of $\mathbf{x}_{i,1:T}$
corresponding to $S$ and $S\cup\left\{ 1,3\right\} $, for each $S$.
Consider one particular value of $S=\left(\left\{ 1,2\right\} \right)$.
The cardinality of $S$, $\left|S\right|=1$.
\begin{itemize}
\item To transform $\mathbf{x}_{i,1:T}$ given $S$, we apply $\varGamma_{1}\left(\mathbf{x}_{i,1:T},S\right)$.
Applying $\varGamma_{1}\left(\mathbf{x}_{i,1:T},S\right)$ transforms
$\mathbf{x}_{i,1:T}$ to $\mathbf{x}_{i,1:T}^{\left(1,S\right)}$
as follows: as per equation (\ref{eq:transformed-X}), for $b'=1$
and tuples $\left\{ 2,2\right\} $ and $\left\{ 1,3\right\} $, set
$x_{i122}^{\left(1,S\right)}=0$ and $x_{i113}^{\left(1,S\right)}=0$.
For all others elements $x_{ib'kt}^{\left(1,S\right)}=x_{ib'kt}$.
Therefore, $\mathbf{x}_{i,1:T}^{\left(1,S\right)}$ = (0,0,0,0,0,0,4,\textbf{\textcolor{red}{0}},0,0,0,10,\textbf{\textcolor{red}{0}},0,0,0,0,0$)'$.
What the transformation has done is to set to 0 the impressions for
brand 1 at ad-position-day combinations $\left\{ 2,2\right\} $ and
$\left\{ 1,3\right\} $, which are the tuples that are not in $S$
at which $i$ saw ads for brand 1 during the $T$ days.
\item To transform $\mathbf{x}_{i,1:T}$ given $S\cup\left\{ 1,3\right\} $,
we apply $\varGamma_{1}\left(\mathbf{x}_{i,1:T},S\cup\left\{ 1,3\right\} \right)$.
Applying $\varGamma_{1}\left(\mathbf{x}_{i,1:T},S\cup\left\{ 1,3\right\} \right)$
transforms $\mathbf{x}_{i,1:T}$ to $\mathbf{x}_{i,1:T}^{\left(1,S\cup\left\{ 1,3\right\} \right)}$
as follows: for $b'=1$ and tuple $\left\{ 2,2\right\} $, set $x_{i122}^{\left(1,S\cup\left\{ 1,3\right\} \right)}=0$.
For all others elements $x_{ib'kt}^{\left(1,S\cup\left\{ 1,3\right\} \right)}=x_{ib'kt}$.
Therefore, $\mathbf{x}_{i,1:T}^{\left(1,S\cup\left\{ 1,3\right\} \right)}$
= (0,0,0,0,0,0,4,\textbf{\textcolor{red}{0}},0,0,0,10,4,0,0,0,0,0$)'$.
What the transformation has done is to set to 0 the impressions for
brand 1 at ad-position-day combination $\left\{ 2,2\right\} $, which
is the only tuple that is not in $S\cup\left\{ 1,3\right\} $ at which
$i$ saw ads for brand 1 during the $T$ days.
\end{itemize}
Evaluating $\sigma\left(.\right)$ at these values now generates the
term in the square brackets in equation (\ref{eq:Shapley-vlaue-Using-response-model}).
Repeating for all possible $S$, and summing per (\ref{eq:Shapley-vlaue-Using-response-model})
gives the Shapley value for brand 1 for tuple $\left\{ k,t\right\} =\left\{ 1,3\right\} $.

\paragraph{An Efficient Algorithm for Fast, Large-Scale Computation}

Exact computation of Shapley values as described above is computationally
intensive. Shapley values have to be calculated separately for each
order. The number of orders can number in the millions on a given
day on an eCommerce platform like \texttt{JD.com}. Additionally, Shapley
values have to be computed for each ad-position-day tuple for each
order. When $\left|\mathcal{N}_{ibT}\right|$ is large, this latter
step also becomes computationally intensive, requiring Monte Carlo
simulation methods to approximate the calculation. 

We seek an implementation that scales to accommodate a large number
of brands and orders, and generates reports in a matter of hours,
which is important for business purposes. Our implementation switches
between exact and approximate solutions for the Shapley values depending
on the cardinality of $\mathcal{N}_{ibT}$, and is implemented in
a MapReduce framework so it runs in a parallel, distributed environment
on a cluster. Algorithm \ref{alg:shap_dist} presents details.
\begin{algorithm}
\caption{Distributed Contribution Computation\label{alg:shap_dist}}
 Input: $List[o\left(i,b,T\right),\mathbf{x}_{i,1:T},\mathbf{p}_{1:T},\mathbf{d}_{i}],\hat{\varOmega},\varGamma,\sigma$ 

Output: the contribution of each ad-position to each brand

\textbf{Function} Map($o\left(i,b,T\right),\mathbf{x}_{i,1:T},\mathbf{p}_{1:T},\mathbf{d}_{i}$)

\begin{algorithmic}[1]
	\State determine $\mathcal{N}_{ibT}$ according to $o\left(i,b,T\right)$ and $ \mathbf{x}_{i,1:T}$
	\State $ Sh[\{ k,t \}] := 0, \{ k,t \} \in \mathcal{N}_{ibT}$
	\If{$ |\mathcal{N}_{ibT}| $ is small enough for exact method}
		\State $\mathcal{P}(\mathcal{N}_{ibT}) := \{ S \mid S \subseteq \mathcal{N}_{ibT} \}$	 	
		\For{$S \in \mathcal{P}(\mathcal{N}_{ibT})$} 	 			 	 		
			\For{$ \{ k,t \} \in \mathcal{N}_{ibT}$} 		 			
				\If{$ \{ k,t \} \in S $} 			 				
					\State $Sh[\{ k,t \}] :=  Sh[\{ k,t \}] + \frac{(|\mathcal{N}_{ibT}|-|S|)!(|S|-1)!}{|\mathcal{N}_{ibT}|!}\sigma(\varGamma_{b}(\mathbf{x}_{i,1:T},S),\mathbf{p}_{1:T},\mathbf{d}_{i};\hat{\varOmega})$
				\Else 			 				
					\State $Sh[\{ k,t \}] :=  Sh[\{ k,t \}] + \frac{(|\mathcal{N}_{ibT}|-|S| - 1)!(|S|)!}{|\mathcal{N}_{ibT}|!}\sigma(\varGamma_{b}(\mathbf{x}_{i,1:T},S),\mathbf{p}_{1:T},\mathbf{d}_{i};\hat{\varOmega})$
				\EndIf 	 	
			\EndFor  	
		\EndFor 
	\Else
		\State $\varPhi := $ the set of permutations of $\mathcal{N}_{ibT}$
		\State $m :=$ number of draws for monte carlo approximation 			
		\State $\varPhi := $ downsample $\varPhi$ to keep $m$ elements
		\For{$\phi  \text{ in } \varPhi$} 	
			\For{$ \{ k,t \} \in \mathcal{N}_{ibT}$} 
				\State $S := \{ \{ k', t' \} \mid \{ k', t' \} \text{ is previous to } \{ k, t \} \text{ in } \phi  \}$
				\State $Sh[\{ k,t \}] :=  Sh[\{ k,t \}] + \frac{1}{|\varPhi|}\sigma(\varGamma_{b}(\mathbf{x}_{i,1:T},S \cup \{ k,t \}),\mathbf{p}_{1:T},\mathbf{d}_{i};\hat{\varOmega})$	
				\State $Sh[\{ k,t \}] :=  Sh[\{ k,t \}]  - \frac{1}{|\varPhi|} \sigma(\varGamma_{b}(\mathbf{x}_{i,1:T},S),\mathbf{p}_{1:T},\mathbf{d}_{i};\hat{\varOmega})$
			\EndFor
		\EndFor
	\EndIf
	\State $SA := sum(Sh)$
	\State $\triangle := \sigma(\varGamma_{b}(\mathbf{x}_{i,1:T},\mathcal{N}_{ibT}),\mathbf{p}_{1:T},\mathbf{d}_{i};\hat{\varOmega}) -  \sigma(\varGamma_{b}(\mathbf{x}_{i,1:T},\varnothing_{b}),\mathbf{p}_{1:T},\mathbf{d}_{i};\hat{\varOmega})$
	\For{$\{ k,t \} \in \mathcal{N}_{ibT}$}	
		\State $Sh[\{ k,t \}] :=  \frac{Sh[\{ k,t \}] * \triangle} { SA}$
	\EndFor
	\State $\widetilde{S}h[k] := 0$ for $k = 1..K$
	\For{$\{ k,t \} \in \mathcal{N}_{ibT}$}	
		\State $\widetilde{S}h[k] :=  \widetilde{S}h[k] + Sh[\{ k,t \}]$
	\EndFor
	\For{$k \in 1..K$}	
		\State $emit((b, k), \widetilde{S}h[k])$
	\EndFor
\end{algorithmic}

\textbf{Function} Reduce($(b,k),List[v]$)

\begin{algorithmic}[1]	 
	\State $SA_k := 0$
	\For{$Sh \text{ in } List[v]$} 	
		\State $SA_k := SA_k + v$
	\EndFor 
	\State $emit((b, k), SA_k)$ 
\end{algorithmic}
\end{algorithm}

\subsubsection{Response Model}

\begin{figure}
\begin{centering}
\centering \caption{Computational Graph for RNN\label{fig:ComputationalGraphRNN}}
\par\end{centering}
\begin{centering}
\includegraphics[scale=0.35]{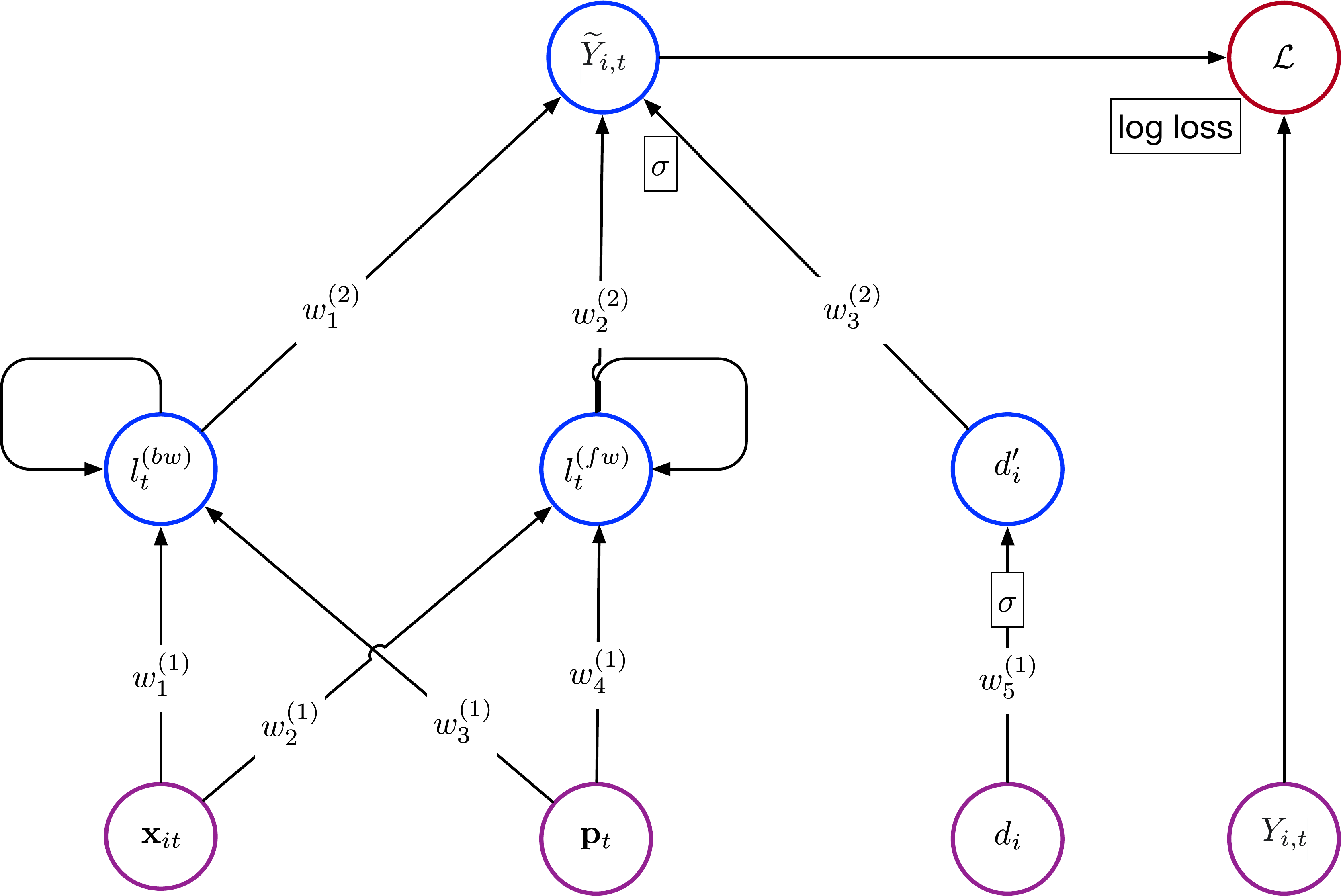}
\par\end{centering}
\begin{spacing}{0.9}
\centering{}\begin{threeparttable}\begin{tablenotes} \item \textit{\scriptsize{}Notes:}{\scriptsize{}
The Figure shows the computational graph for the RNN model for ad-response.
Though the model training is done simultaneously across all brands,
the picture is drawn only for one brand. The input vector of ad-impressions,
$\mathbf{x_{it}}$, and the input vector of price-indexes, $\mathbf{p}_{t}$,
are fed through an LSTM layer with recurrence. The user characteristics,
$\mathbf{d}_{i}$,} {\scriptsize{}are processed through a separate,
fully-connected layer. The outputs from the LSTM cells and the fully-connected
layer jointly impact the predicted outcome $\tilde{Y}_{it}$. Combining
this with the observed outcome,$Y_{it}$, we obtain the log-likelihood
$\mathcal{L}$, which forms the loss function for the model.}\end{tablenotes}
\end{threeparttable} 
\end{spacing}
\end{figure}

The purpose of the response model is to provide a data-driven way
to estimate the marginal effects in equation (\ref{eq:Shapley-value-k-t}).
The architecture of the RNN is presented in Figure (\ref{fig:ComputationalGraphRNN}).
Though the model training is done simultaneously across all brands,
the picture is drawn only for one brand. The input vector of ad-impressions,
$\mathbf{x_{it}}$, and the input vector of price-indexes, $\mathbf{p}_{t}$,
are fed through an LSTM layer with recurrence. The user characteristics,
$\mathbf{d}_{i}$, are processed through a separate, fully-connected
layer. The outputs from the LSTM cells and the fully-connected layer
jointly impact the predicted outcome $\tilde{Y}_{it}$. Combining
this with the observed outcome,$Y_{it}$, we obtain the log-likelihood
$\mathcal{L}$, which forms the loss function for the model. The RNN
finds a set of parameters or weights that maximizes the log-likelihood.

As noted before, we utilize a bi-directional formulation in which
we allow for a hidden layer with backward recurrence, augmented with
a hidden layer with forward recurrence. The layer with forward recurrence
serves as a semi-parametric summary of future activity that is helpful
to predict current actions. This is shown in Figure (\ref{fig:ComputationalGraphRNN})
where the superscript ``$\textrm{fw}$'' indicates forward recurrence
and ``$\textrm{bw}$'' indicates backward recurrence. The use of
``future'' ad-impressions for predicting current behavior in the
response model requires some elaboration when the model is used to
compute the ``causal'' or marginal effects in the Shapley Values.
Note that the causal effects the response model has to deliver for
computing the Shapley values for a user, are always differences in
the user's predicted probabilities of purchase in a period $T$, under
different \textit{retrospective} (pre$-T$) counterfactual ad-impression
sequences. This means the use of future ad-impressions to predict
behavior in the bi-directional model poses no conceptual difficulty
in developing the causal effects required for computing Shapley Values.

The model is implemented in \texttt{TensorFlow}. We use the non-peephole
based implementation of the LSTM \citep{HochreiterSchmidhuber97};
regularized via dropout with probability $0.75$; and optimized via
stochastic gradient descent using the \texttt{Adam Optimizer} \citep{KingmaBa14}.
We initialize the forward and backward LSTM cells to 0; the LSTM weights
orthogonally with multiplicative factor = 1.0; and the other parameters
using truncated normal draws. The model training is stopped with the
error in a validation dataset stabilizes.

\section{Experiments and Application to Cell-phone Product Category}

We present an application of the model using individual-level data
on ad-exposures and purchases from the cell-phone product category
on \texttt{JD.com} during a 15-day window in 2017. We first present
some model-free evidence documenting the quantitative relevance in
our data of some of the considerations outlined in the introduction
for a good response model. Then we show present model performance
metrics and results. 

\paragraph{Data and Summary Statistics}

To create the training data, we sample users who saw during the 15-day
window, at least one ad-impression related to a product in the cell-phone
product category sold on \texttt{JD.com}. Within this overall sample,
we define the\textit{ positive sample} as the set of users who purchased
a product of any brand in the cell-phone category during the time
window. We define the\textit{ negative sample} as the set of users
in the overall sample who did not purchase any product in the cell-phone
category during the 15-day time window. Table \ref{tab:Summary-Statistics-of-TD}
provides summary statistics of the training dataset. There are roughly
75M users, 3.4M orders, and 7B ad-impressions. There are 301 ad-positions.
We aggregate brands to 31. Table (\ref{tab:Market-Share-ofUnits})
shows market shares on the basis of units sold and revenue generated.
\texttt{Huawei} is the largest brand, followed by \texttt{Apple} (2nd
largest in terms of revenue), \texttt{Xiaomi}, \texttt{Meizhu}, \texttt{Vivo}
and others.
\begin{table}
\caption{Summary Statistics of Training Data\label{tab:Summary-Statistics-of-TD}}

\medskip{}

\begin{centering}
\begin{tabular}{r|r}
\hline 
{\footnotesize{}Number of users in Overall Sample} & {\scriptsize{}75,768,508}\tabularnewline
{\footnotesize{}Number of users in Positive Sample} & {\scriptsize{}2,100,687}\tabularnewline
{\footnotesize{}Number of users in Negative Sample} & {\scriptsize{}73,667,821}\tabularnewline
{\footnotesize{}Number of ad-impressions in }product category{\footnotesize{}
over 15 days} & {\scriptsize{}7,153,997,856}\tabularnewline
{\footnotesize{}Number of orders made in }product category{\footnotesize{}
over 15 days} & {\scriptsize{}3,477,621}\tabularnewline
{\footnotesize{}Number of orders made on day $T=15$ } & {\scriptsize{}175,937}\tabularnewline
\hline 
{\footnotesize{}Number of brands ($B$)} & {\footnotesize{}31}\tabularnewline
{\footnotesize{}Number of ad-positions ($K$)} & {\footnotesize{}301}\tabularnewline
\hline 
\end{tabular}
\par\end{centering}
\begin{spacing}{0.9}
\centering{}\begin{threeparttable}\begin{tablenotes} \item \textit{\footnotesize{}Notes:}{\scriptsize{}
Descriptive statistics of training dataset, which comprises individual-level
data on ad-exposures and purchases from the cell-phone category on
}\texttt{\scriptsize{}JD.com}{\scriptsize{} during a 15-day window
in 2017. The}\textit{\scriptsize{} positive sample}{\scriptsize{}
is the set of users who purchased a product of any brand in the cell-phone
category during the time window. The}\textit{\scriptsize{} negative
sample}{\scriptsize{} is the set of users in the overall sample who
did not purchase any product in the cell-phone category during the
15-day time window.} \end{tablenotes} \end{threeparttable} 
\end{spacing}
\end{table}
\begin{table}
\caption{Market Shares in Category by Brand\label{tab:Market-Share-ofUnits}}

\medskip{}

\begin{centering}
\begin{tabular}{c|c|c|c|c|c|c}
\hline 
\multirow{3}{*}{} & \multirow{3}{*}{Huawei} & \multirow{3}{*}{Xiaomi} & \multirow{3}{*}{Apple} & \multirow{3}{*}{Meizhu} & \multirow{3}{*}{Vivo} & \multirow{3}{*}{Others}\tabularnewline
 &  &  &  &  &  & \tabularnewline
 &  &  &  &  &  & \tabularnewline
\hline 
{\footnotesize{}By Units Sold} & {\scriptsize{}29.5\% } & {\scriptsize{}25.30\%} & {\scriptsize{}8.40\%} & {\scriptsize{}6.20\%} & {\scriptsize{}3.20\%} & {\scriptsize{}27.40\%}\tabularnewline
\hline 
{\footnotesize{}By RMB Sold} & {\scriptsize{}27.2\%} & {\scriptsize{}18.4\%} & {\scriptsize{}24.7\%} & {\scriptsize{}4.2\%} & {\scriptsize{}5.3\%} & {\scriptsize{}20.2\%}\tabularnewline
\hline 
\end{tabular}
\par\end{centering}
\begin{spacing}{0.9}
\centering{}\begin{threeparttable}\begin{tablenotes} \item \textit{\footnotesize{}Notes:}{\scriptsize{}
Brand-level market shares in the training data based on units sold
and by money spent (RMB) in the cell-phone category on }\texttt{\scriptsize{}JD.com}{\scriptsize{}
during a 15-day window in 2017.}\end{tablenotes} \end{threeparttable} 
\end{spacing}
\end{table}
\begin{table}
\caption{Summary Statistics of Ad-Exposures by Brand\label{tab:Summary-Statistics-ofAdExposures}}
\medskip{}

\centering{}%
\begin{tabular}{r|cc|cc}
\hline 
\noalign{\vskip0.25cm}
Brand & \multicolumn{2}{c|}{Positive Sample} & \multicolumn{2}{c}{Negative Sample}\tabularnewline[0.25cm]
\hline 
 & Mean & SD & Mean & SD\tabularnewline
\hline 
Huawei & {\footnotesize{}201.6} & {\footnotesize{}328.6} & {\footnotesize{}34.7} & {\footnotesize{}144.9}\tabularnewline
Xiaomi & {\footnotesize{}220.3} & {\footnotesize{}371.6} & {\footnotesize{}26.3} & {\footnotesize{}111.7}\tabularnewline
Apple & {\footnotesize{}147.2} & {\footnotesize{}253.1} & {\footnotesize{}16.5} & {\footnotesize{}63.3}\tabularnewline
Meizhu & {\footnotesize{}173.4} & {\footnotesize{}309.9} & {\footnotesize{}14.3} & {\footnotesize{}65.8}\tabularnewline
Vivo & {\footnotesize{}82.4} & {\footnotesize{}147.9} & {\footnotesize{}15.2} & {\footnotesize{}46.6}\tabularnewline
Others & {\footnotesize{}97.6} & {\footnotesize{}197.7} & {\footnotesize{}8.1} & {\footnotesize{}43.2}\tabularnewline
\hline 
\end{tabular}
\end{table}

\paragraph{Motivating Patterns in Data}

Table (\ref{tab:Summary-Statistics-ofAdExposures}) shows summary
statistics of ad-exposures split by brand separately for the positive
sample and the negative sample. Ad-exposures are seen to much higher
in the positive sample. For example, mean exposures in the positive
sample (over a 15-day windows) are 737\% ($\frac{220.3-26.3}{26.3}$)
higher for \texttt{Xiaomi}, and 792\% ($\frac{147.2-16.5}{16.5}$)
for \texttt{Apple}. While some of this can be driven by targeting
by the brands of users more likely to buy their products, the large
differences across the positive and negative samples by brand suggest
that the sequence of their ad-exposures over the 15-day window matter
in explaining purchases on day-15.

\begin{sidewaysfigure}[ph]
\begin{centering}
\centering \caption{Intensity of Advertising Exposure Matters for Conversion\label{fig:Intensity-of-Advertising}}
\subfloat[Apple\label{fig:Intensity-of-Advertising:Apple}]{
\centering{}\includegraphics[scale=0.7]{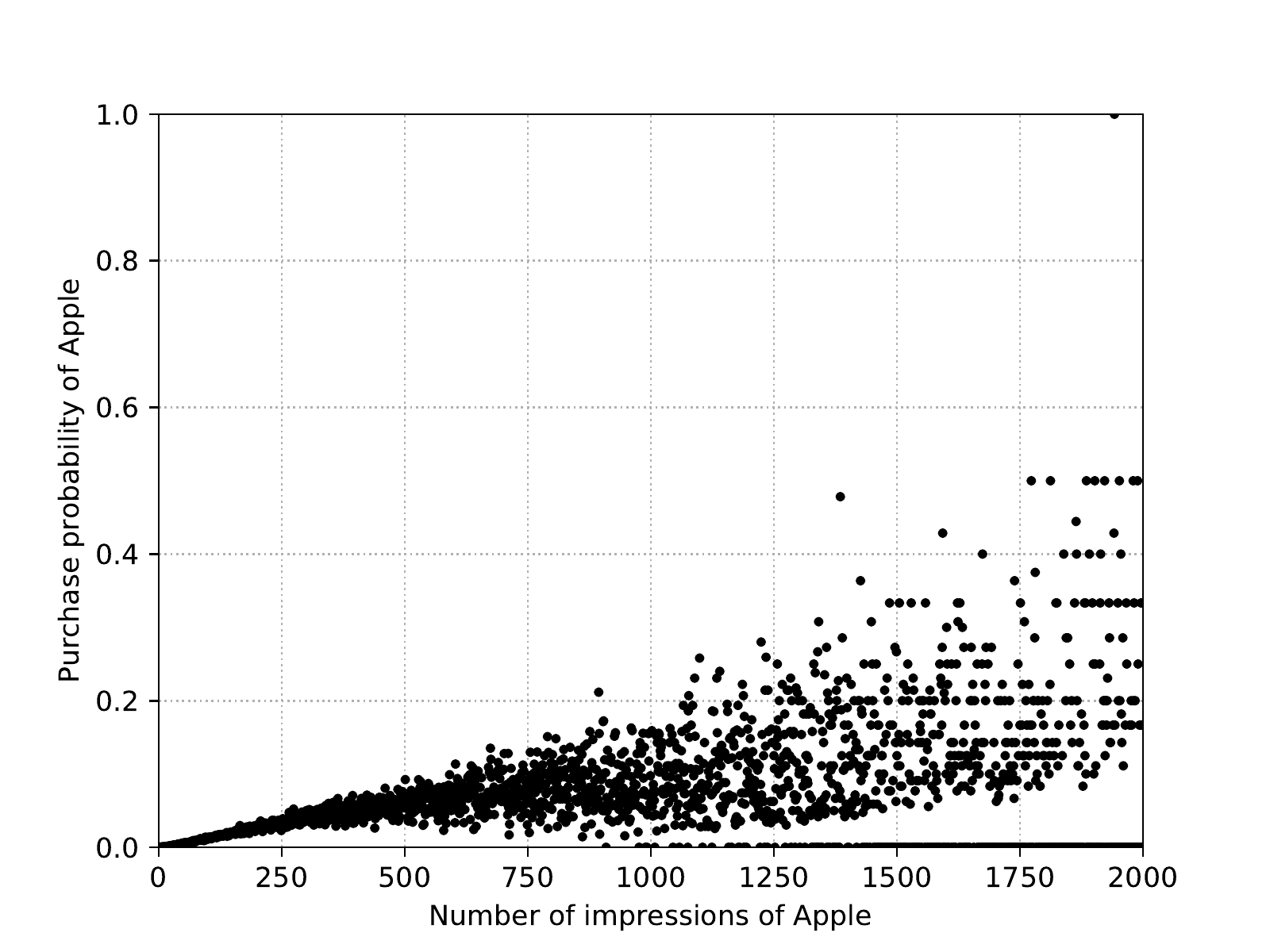}}\subfloat[Xiaomi\label{fig:Intensity-of-Advertising:Xiaomi}]{
\centering{}\includegraphics[scale=0.7]{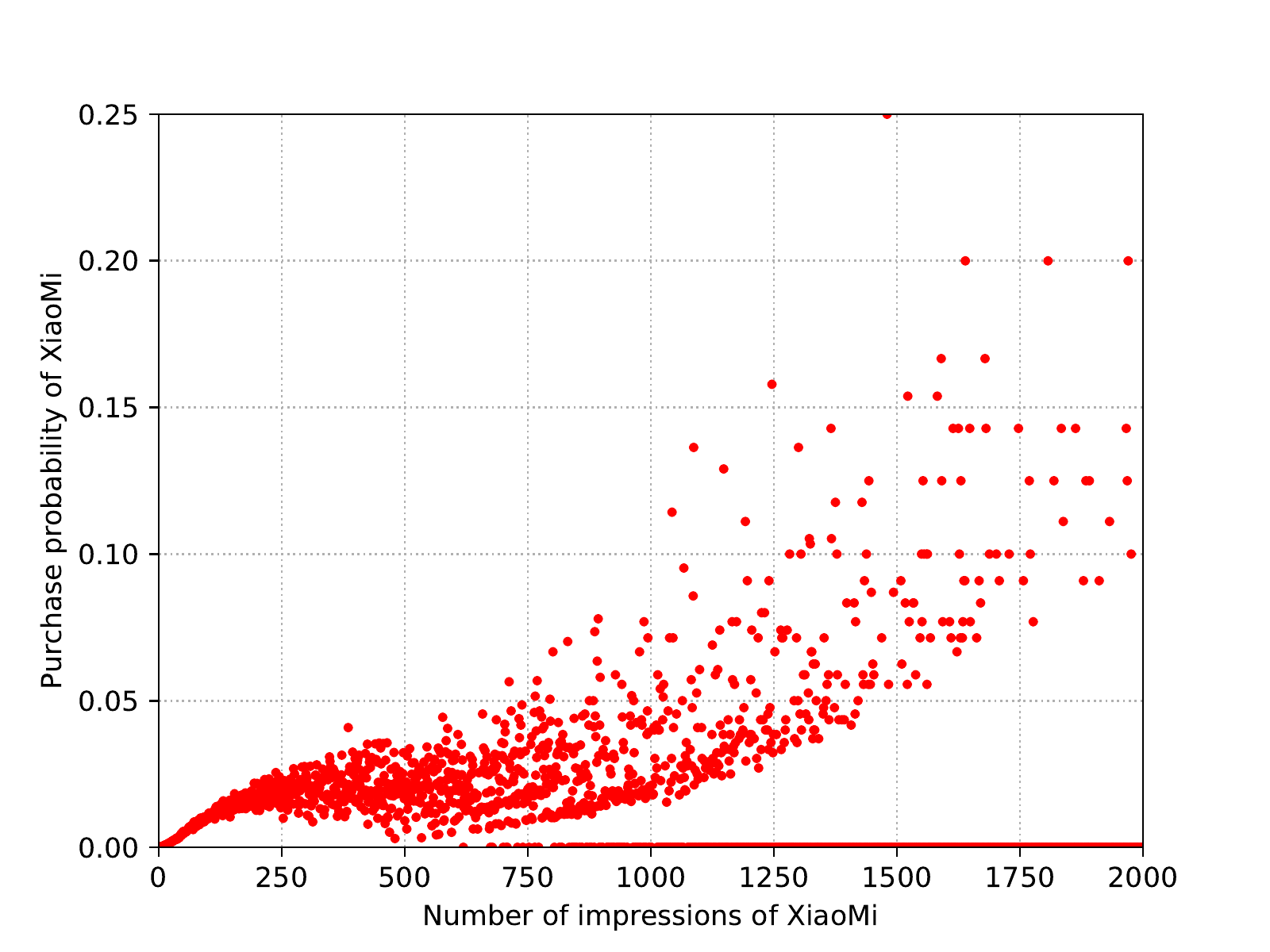}}
\par\end{centering}
\begin{spacing}{0.9}
\centering{}\begin{threeparttable}\begin{tablenotes} \item \textit{\footnotesize{}Notes:}{\scriptsize{}
The Figure shows a plot of the probability of purchase of a brand
on day $15$ as a function of the number of impressions of ads for
}\texttt{\scriptsize{}Apple}{\scriptsize{} (left panel) or }\texttt{\scriptsize{}Xiaomi}{\scriptsize{}
(right panel) seen by the user in the past 15 days. To construct the
plot, define the negative sample as the set of users in the overall
sample who did not purchase any product in the cell-phone category
during the 15-day time window. At each value of the $x$-axis (number
of own-brand ad-impressions), the probability of purchase of the brand
on the $y$-axis is computed as the number of users who bought the
brand's products on day-15, divided by the total numbers of users
in the negative sample. The $x$-axis is capped at 2,000 impressions
to account for bots, crawlers, non-individual buyers etc.} \end{tablenotes}
\end{threeparttable} 
\end{spacing}
\end{sidewaysfigure}

\begin{sidewaysfigure}[ph]
\begin{centering}
\centering \caption{Timing of Advertising Exposure Matters for Conversion\label{fig:Timing-of-Advertising}}
\subfloat[Apple\label{fig:Timing-of-Advertising:Apple}]{
\centering{}\includegraphics[scale=0.7]{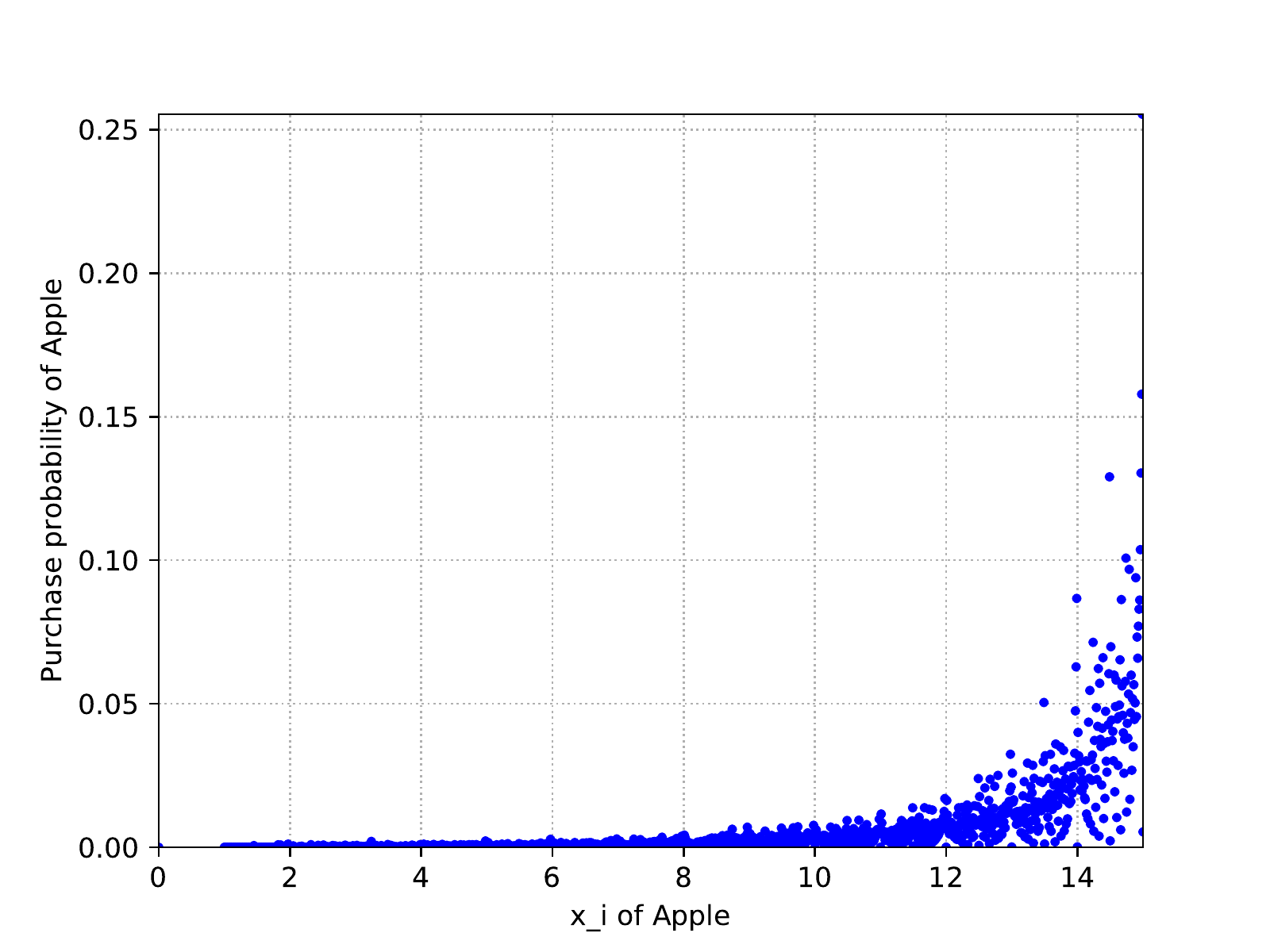}}\subfloat[Xiaomi\label{fig:Timing-of-Advertising:Xiaomi}]{
\centering{}\includegraphics[scale=0.7]{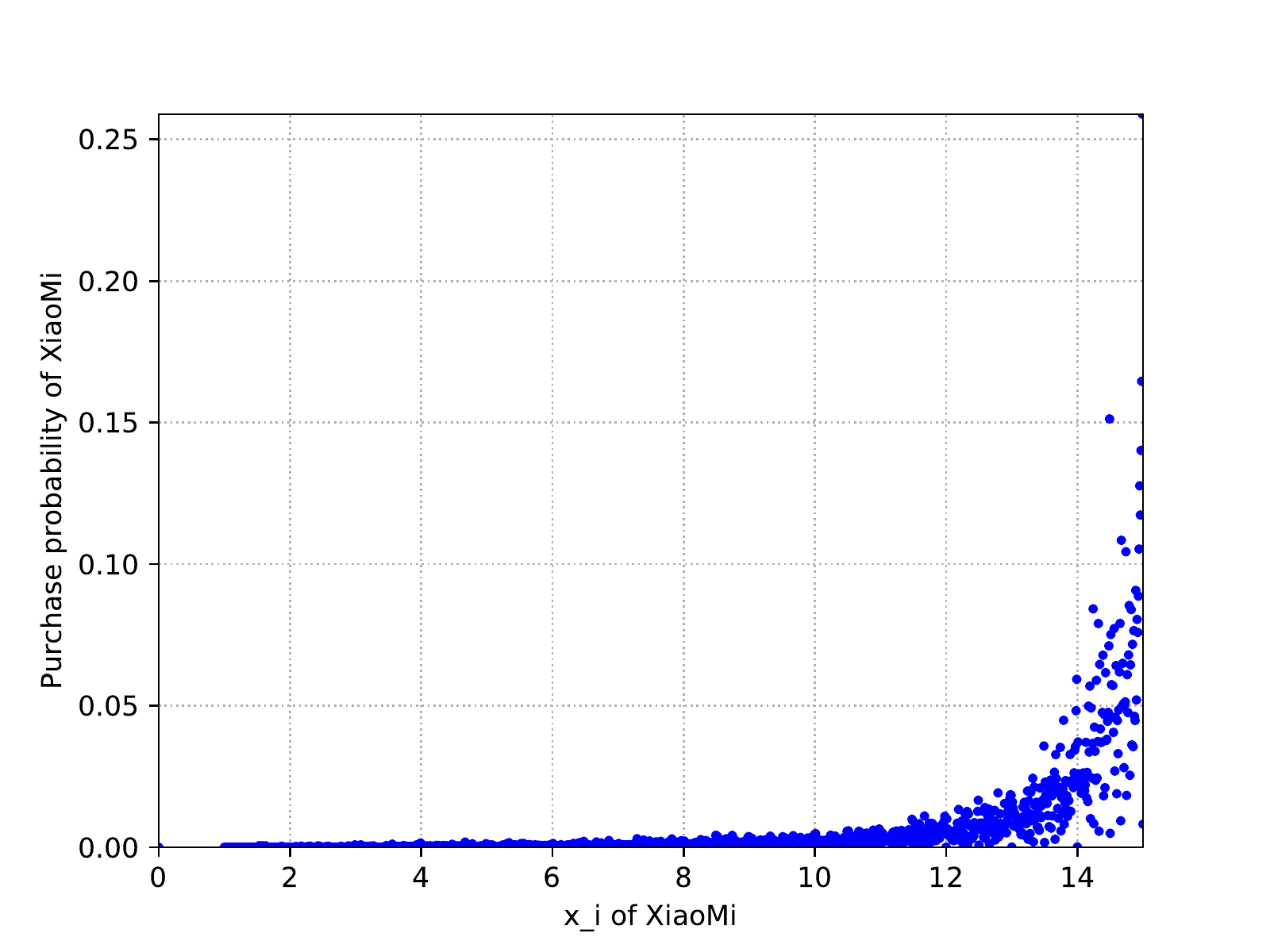}}
\par\end{centering}
\begin{spacing}{0.9}
\centering{}\begin{threeparttable}\begin{tablenotes} \item \textit{\footnotesize{}Notes:}{\scriptsize{}
The Figure shows a plot of the probability of purchase of a brand
on day-15 as a function of the days since the user saw impressions
of ads for that brand. Plots are presented separately for }\texttt{\scriptsize{}Apple}{\scriptsize{}
(left panel) and }\texttt{\scriptsize{}Xiaomi}{\scriptsize{} (right
panel). To construct the plot, define the negative sample as the set
of users in the overall sample who did not purchase any product in
the cell-phone category during the 15-day time window. At each value
of the $x$-axis, the probability of purchase of the brand on the
$y$-axis is computed as the number of users who bought the brand's
products on day-15, divided by the total numbers of users in the negative
sample. The $x$-axis represents the average days elapsed since the
user saw ad-impressions of that brand, obtained by weighting the day
of the ad-impression by the number of impressions over the 15 days.
Specifically, for each user $i$, the $x$-value for user $i$ for
brand $b$, $x_{ib}$ is computed as $\sum_{t=1}^{15}t\times\frac{n_{ibt}}{\sum_{t=1}^{15}n_{ibt}}$,
where $n_{ibt}$ is the number of impressions of ads of brand $b$seen
by user $i$ on day $t$, $t=1,..,15$. }\end{tablenotes} \end{threeparttable} 
\end{spacing}
\end{sidewaysfigure}

Figure (\ref{fig:Intensity-of-Advertising}) shows plots of the probability
of purchase of a brand on day $15$ as a function of the number of
impressions of ads for \texttt{Apple} (left panel) or \texttt{Xiaomi}
(right panel) seen by the user in the past 15 days. There is evidence
of a robust positive association, suggesting that the intensity of
own-advertising exposure matters for conversion. Figure (\ref{fig:Timing-of-Advertising})
shows the probability of purchase of a brand on day-15 as a function
of the days since the user saw impressions of ads for that brand.
Plots are presented separately for \texttt{Apple} (left panel) and
\texttt{Xiaomi} (right panel). To represent exposure timing, we represent
on the $x$-axis, the average days elapsed since a user saw ad-impressions
of that brand, obtained by weighting the day of the ad-impression
by the number of impressions over the 15 days. Specifically, for each
user $i$, the $x$-value for user $i$ for brand $b$, $x_{ib}$
is computed as $\sum_{t=1}^{15}t\times\frac{n_{ibt}}{\sum_{t=1}^{15}n_{ibt}}$,
where $n_{ibt}$ is the number of impressions of ads of brand $b$
seen by user $i$ on day $t$, $t=1,..,15$. The response curve is
U-shaped, with more recent exposures associated with higher purchase
probability of the brand on day 15. This suggests decay in the ad-response.
The plot also shows the effect decays close to zero in 15 days, providing
some data-based justification for using this cutoff.

\begin{sidewaysfigure}[ph]
\begin{centering}
\centering \caption{Exposure to Competitor Advertising Exposure Matters for Conversion\label{fig:Exposure-to-CompetitorAds}}
\subfloat[Apple\label{fig:ExposuretoCompetitorAdvertising:Apple}]{
\centering{}\includegraphics[scale=0.7]{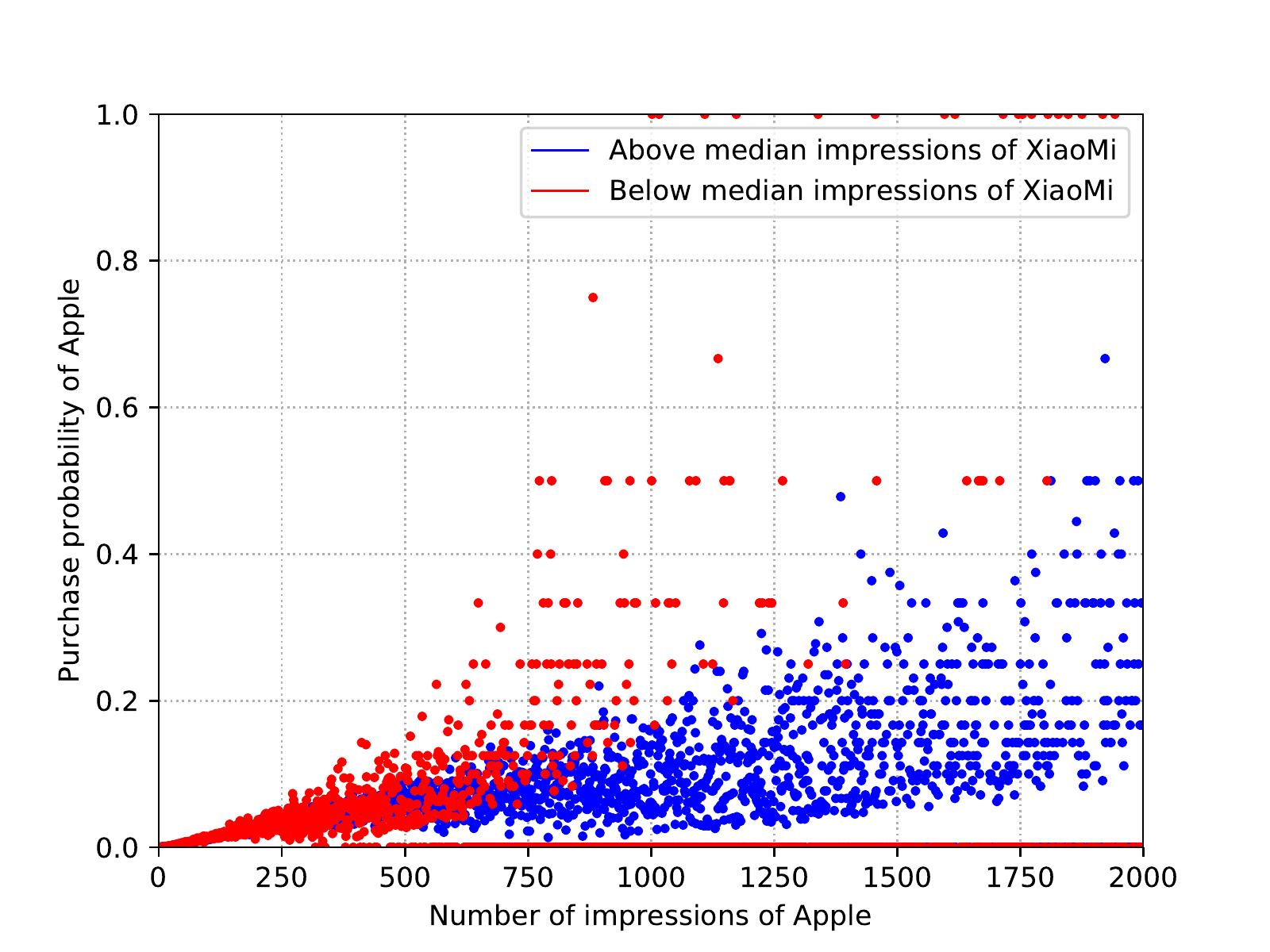}}\subfloat[Xiaomi\label{fig:ExposuretoCompetitorAdvertising:Xiaomi}]{
\centering{}\includegraphics[scale=0.7]{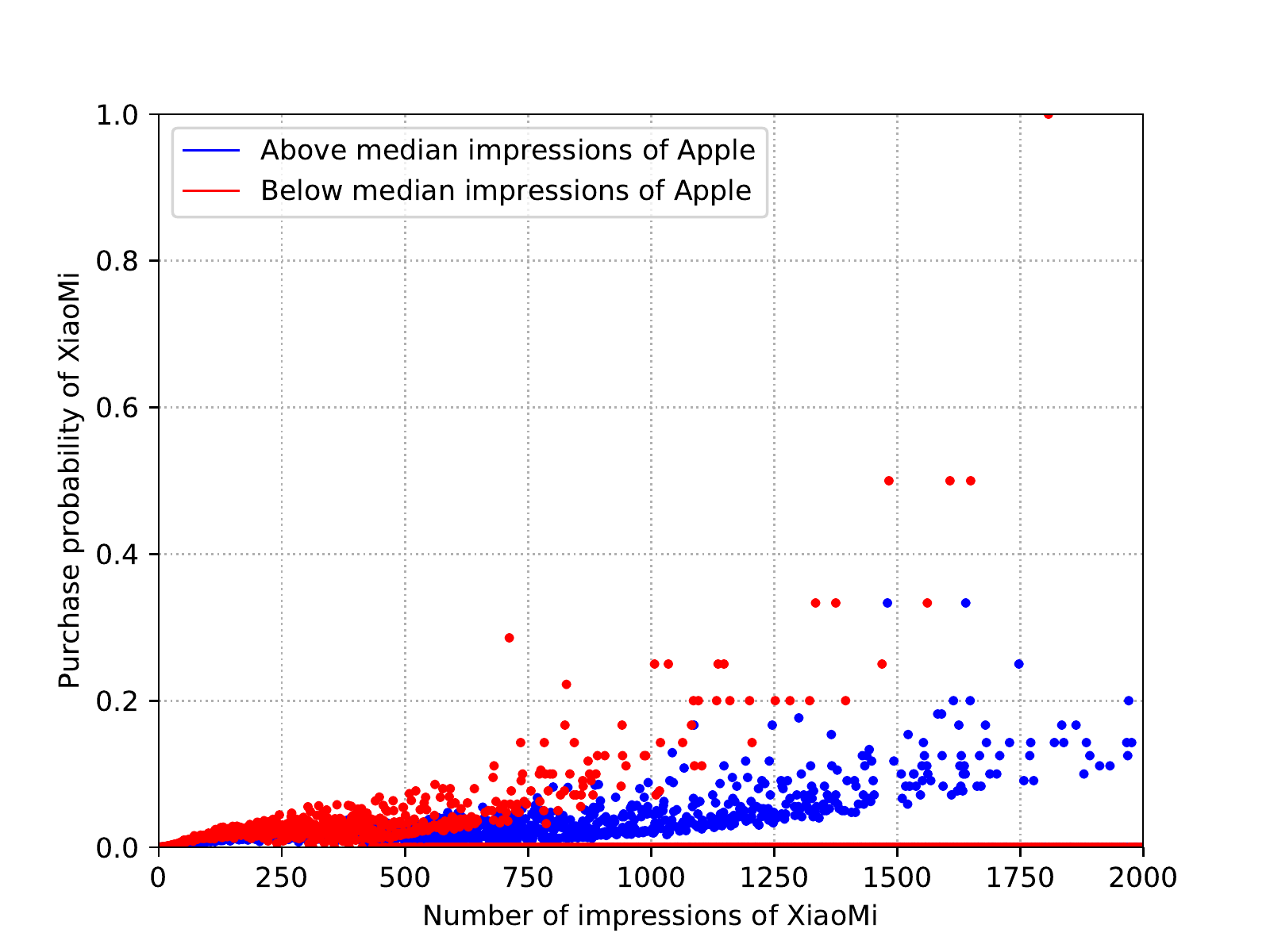}}
\par\end{centering}
\begin{spacing}{0.9}
\centering{}\begin{threeparttable}\begin{tablenotes} \item \textit{\footnotesize{}Notes:}{\scriptsize{}
The Figure shows a plot of the probability of purchase of a brand
on day-15 as a function of the number of impressions of ads for }\texttt{\scriptsize{}Apple}{\scriptsize{}
(left panel) seen by the user over the 15 day window. To construct
the plot, define the negative sample as the set of users in the overall
sample who did not purchase any product in the cell-phone category
during the 15-day time window. At each value of the $x$-axis (number
of own-brand ad-impressions), the probability of purchase of the brand
on the $y$-axis is computed as the number of users who bought the
brand's products on day-15, divided by the total numbers of users
in the negative sample. The $x$-axis is capped at 2,000 impressions
to account for bots, crawlers, non-individual buyers etc. The probability
of purchase of }\texttt{\scriptsize{}Apple}{\scriptsize{} is plotted
separately by a median split of the number of impressions seen of
of }\texttt{\scriptsize{}Xiaomi}{\scriptsize{} ads. The blue dots
depict the probabilities for those who saw more than the median impressions
of }\texttt{\scriptsize{}Xiaomi}{\scriptsize{} ads over the 15-day
window, and the red dots depict the probabilities for those who saw
less than the median impressions of }\texttt{\scriptsize{}Xiaomi}{\scriptsize{}
ads over the 15-day window. There is evidence for separation: the
association of purchase probabilities with }\texttt{\scriptsize{}Apple}{\scriptsize{}
ad-impressions is steeper for those who saw less than the median impressions
of }\texttt{\scriptsize{}Xiaomi}{\scriptsize{} ads over the 15-day
window. The right panel depicts analogous plot for }\texttt{\scriptsize{}Xiaomi}{\scriptsize{}.}\end{tablenotes}
\end{threeparttable} 
\end{spacing}
\end{sidewaysfigure}

Figure (\ref{fig:Exposure-to-CompetitorAds}) shows a plot of the
probability of purchase of a brand on day-15 as a function of the
number of impressions of ads for \texttt{Apple} (left panel) seen
by the user over the 15 day window. The probability of purchase of
\texttt{Apple} is plotted separately by a median split of the number
of impressions seen of \texttt{Xiaomi} ads. The blue dots depict the
probabilities for those who saw more than the median impressions of
\texttt{Xiaomi} ads over the 15-day window, and the red dots depict
the probabilities for those who saw less than the median impressions
of \texttt{Xiaomi} ads over the 15-day window. There is evidence for
separation: the association of purchase probabilities with \texttt{Apple}
ad-impressions is steeper for those who saw less than the median impressions
of \texttt{Xiaomi} ads over the 15-day window. The right panel depicts
the analogous plot for \texttt{Xiaomi}. The pattern is similar, suggesting
the importance of allowing for competitor ads to matter in affecting
purchases.

\begin{figure}[H]
\begin{centering}
\centering \caption{Comparison of Marginal Effects from a Linear Response Model With and
Without User Fixed Effects\label{fig:Comparison-of-Marginal-Effects-FE}}
\par\end{centering}
\begin{centering}
\includegraphics[scale=0.75]{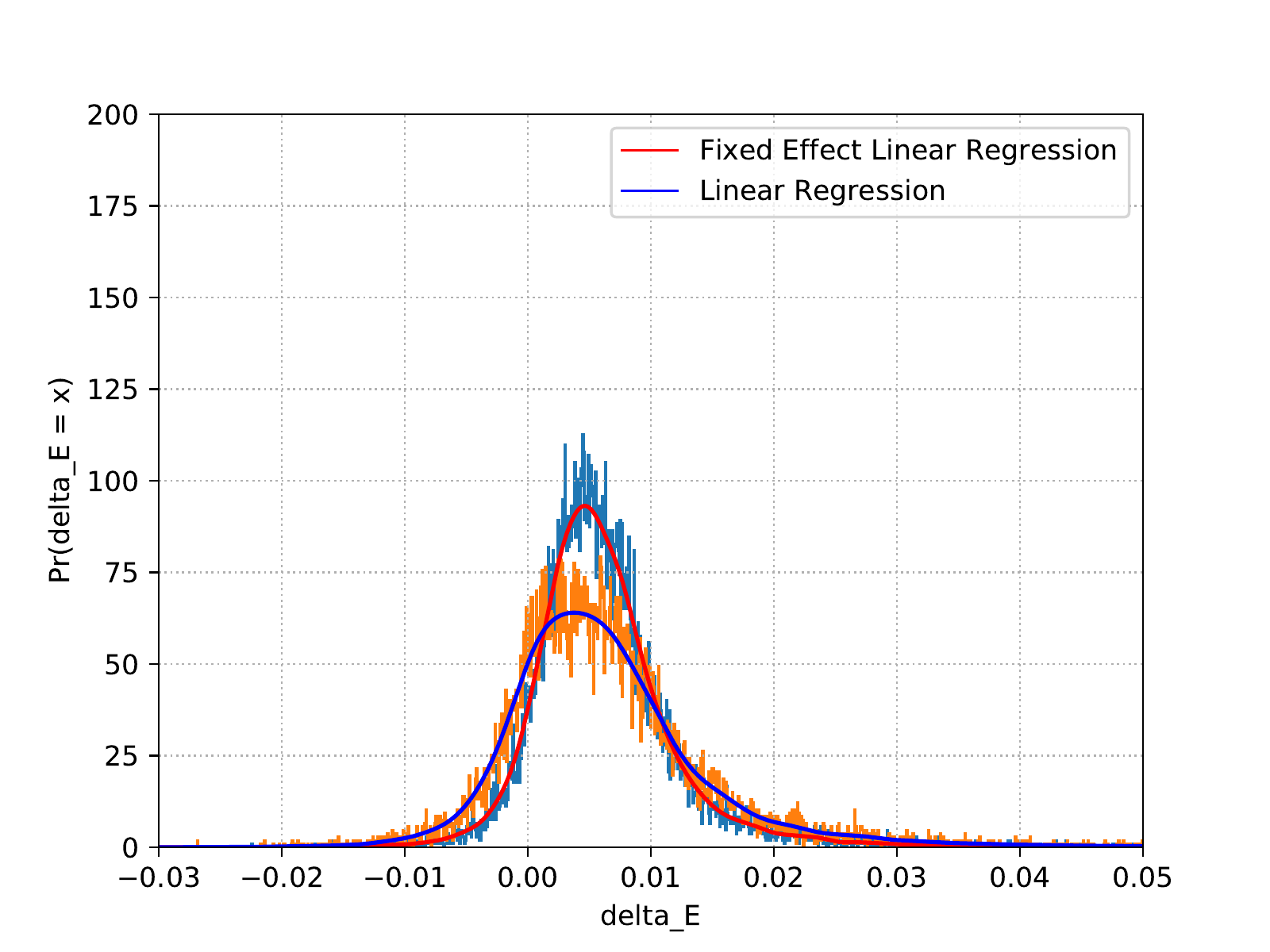}
\par\end{centering}
\begin{spacing}{0.9}
\centering{}\begin{threeparttable}\begin{tablenotes} \item \textit{\scriptsize{}Notes:}{\scriptsize{}
Histogram of estimated marginal effects for a search ad-position using
linear models with and without ``fixed-effects.'' Both models put
significant probability mass on the positive support, but fewer of
the marginal effects are negative under the fixed-effects model.}\end{tablenotes}
\end{threeparttable} 
\end{spacing}
\end{figure}

Finally, in Figure (\ref{fig:Comparison-of-Marginal-Effects-FE}),
we assess the importance of including user characteristics in the
model as a control for user heterogeneity and selection into ad-exposure.
We do this informally by comparing the marginal effects for a search
ad-position for linear models with and without ``fixed-effects.''
Comparing to a model without, the model with fixed-effects allows
flexible user heterogeneity on the intercept. We choose a search ad-position
because it receives some of the highest ad-impressions in the data,
and also because the issue of selection is likely to be severe in
the case of search ads (i.e., those who like a brand are more likely
to search for it and see the search ads, while being more likely to
buy the product without the exposure). If we find that the predicted
marginal effects are ``more reasonable'' under the fixed-effects
model compared to the base model, that provides some evidence for
the value of including user heterogeneity. Past ``A/B'' testing
at \texttt{JD.com} has shown that the search ad-position produces
positive marginal lift across many historical campaigns. So, we use
the extent to which the predicted results are positive as a metric
of reasonableness.

To do this, we let $y_{i}$ be an indicator of whether user $i$ bought
a product of a given brand on day 15 (to economize on notation, the
index $b$ for brand is suppressed). We first train two models for
$y_{i}$: (1) a linear model $y_{i}=\alpha+\sum_{k=1}^{K}\beta_{k}x_{ik}$,
where $x_{ik}$ is the number of impressions seen by user $i$ of
that brand at ad-position $k$ over the 15-day window; and, (2) a
linear fixed-effects model $y_{i}=\alpha_{i}+\sum_{k=1}^{K}\beta_{k}x_{ik}$,
which is the same as the previous model, except the intercept $\alpha$
is $i$-specific. Once the two models are trained, we store them in
memory. The predictions from the two models are denoted $\hat{y}_{i}^{L}$
and $\hat{y}_{i}^{L-FE}$ respectively. Our goal is to compare predictions
between the two models for a focal search ad-position, denoted by
$\tilde{k}$. The marginal value of seeing ads at position $\tilde{k}$
depends on the sequence of ads preceding it, so in order to do the
comparison between models, we need to pick a set of sequences on which
to base the comparison. We would like to pick the sequences that occur
most in the data and for which we have enough data so that we can
assess the comparison with a reasonable amount of precision. With
these considerations, we filter to the sub-set of users in the data
who saw ad-impressions at least 10 ad-positions over the 15-day window
(10 is the median across users in the data). For each $i$ in this
sub-set, we let $S_{i}$ denote the sequence of ad-positions that
$i$ saw over the 15-day window (so, by construction, $\left\Vert S_{i}\right\Vert =10$).
Define by $S$, a 9-element permutation from the set of ad-positions
excluding $\tilde{k}$, i.e. a 9-element permutation from the set
$\left(1,..,K\right)\left\backslash \left\{ \tilde{k}\right\} \right.$.
We then estimate the marginal value to seeing ads at position $\tilde{k}$
when it occurs as the $10^{th}$ in the sequence, given that the first
9 ad-positions seen is $S$ as, 
\[
\triangle^{L}\left(S,\tilde{k}\right)=\mathbb{E}_{i}\left[\hat{y}_{i}^{L}|S_{i}\equiv\left(S\cup\tilde{k}\right)\right]-\mathbb{E}_{i}\left[\hat{y}_{i}^{L}|S_{i}\equiv\left(S\cup k\right),k\neq\tilde{k}\right]
\]
The first term is obtained by averaging the predicted $\hat{y}_{i}^{L}$
using the observed ad-impressions for all users $i$ that have the
sequence $S_{i}\equiv\left(S\cup\tilde{k}\right)$, and the second
term is obtained by averaging the predicted $\hat{y}_{i}^{L}$ using
the observed impressions for all users $i$ that have the sequence
$S_{i}\equiv\left(S\cup k\right),k\neq\tilde{k}$. For each 9-position
sequence $S,$ $\triangle^{L}\left(S,\tilde{k}\right)$ thus provides
an estimate using the linear model, of the incremental benefit of
seeing ads at position $\tilde{k}$ next, rather than at position
$k\neq\tilde{k}$ next. We do the same thing using the fixed effects
model, to compute analogously $\triangle^{L-FE}\left(S,\tilde{k}\right)$
for each $S.$ Then, we plot a histogram of the distribution of $\triangle^{L}\left(S,\tilde{k}\right)$
and $\triangle^{L-FE}\left(S,\tilde{k}\right)$ across $S$ for ad-position
$\tilde{k}$. This is shown in Figure (\ref{fig:Comparison-of-Marginal-Effects-FE}). 

Looking at Figure (\ref{fig:Comparison-of-Marginal-Effects-FE}),
we see that both models put significant probability mass on the positive
support, but fewer of the marginal effects are negative under the
fixed-effects model. We take this as supportive evidence that including
controls for selection is important to generate reasonable measures
of ad-effectiveness, apart from allowing effects to be estimated separately
by user segment.

\paragraph*{Model Performance}

Table \ref{tab:RNNModelPerformanceMetricsbyBrand} shows the accuracy,
recall and precision of the model. At the brand level, the model has
precision ranging from 69\%$-$76\% and recall ranging from 12\%$-$34\%
for the top brands; showing it fits the data well.
\begin{table}
\caption{RNN Model Performance Metrics by Brand\label{tab:RNNModelPerformanceMetricsbyBrand}}
\medskip{}

\centering{}%
\begin{tabular}{r|ccc}
\hline 
\noalign{\vskip0.25cm}
Brand & \multicolumn{3}{c}{Metric}\tabularnewline[0.25cm]
\hline 
 & Accuracy & Recall & Precision\tabularnewline
\hline 
Huawei & {\footnotesize{}0.997} & {\footnotesize{}0.340} & {\footnotesize{}0.714}\tabularnewline
Xiaomi & {\footnotesize{}0.997} & {\footnotesize{}0.308} & {\footnotesize{}0.705}\tabularnewline
Apple & {\footnotesize{}0.998} & {\footnotesize{}0.218} & {\footnotesize{}0.696}\tabularnewline
Meizhu & {\footnotesize{}0.999} & {\footnotesize{}0.186} & {\footnotesize{}0.727}\tabularnewline
Vivo & {\footnotesize{}0.999} & {\footnotesize{}0.119} & {\footnotesize{}0.762}\tabularnewline
Others & {\footnotesize{}1.000} & {\footnotesize{}0.095} & {\footnotesize{}0.726}\tabularnewline
\hline 
\end{tabular}
\end{table}

\begin{figure}
\begin{centering}
\centering \caption{Performance of Bi-directional RNN Relative to Benchmark Specifications\label{fig:Performance-of-BidirectionalRNNBenchmarks}}
\par\end{centering}
\begin{centering}
\subfloat[Accuracy\label{fig:accuracy_validation}]{
\centering{}\includegraphics[scale=0.53]{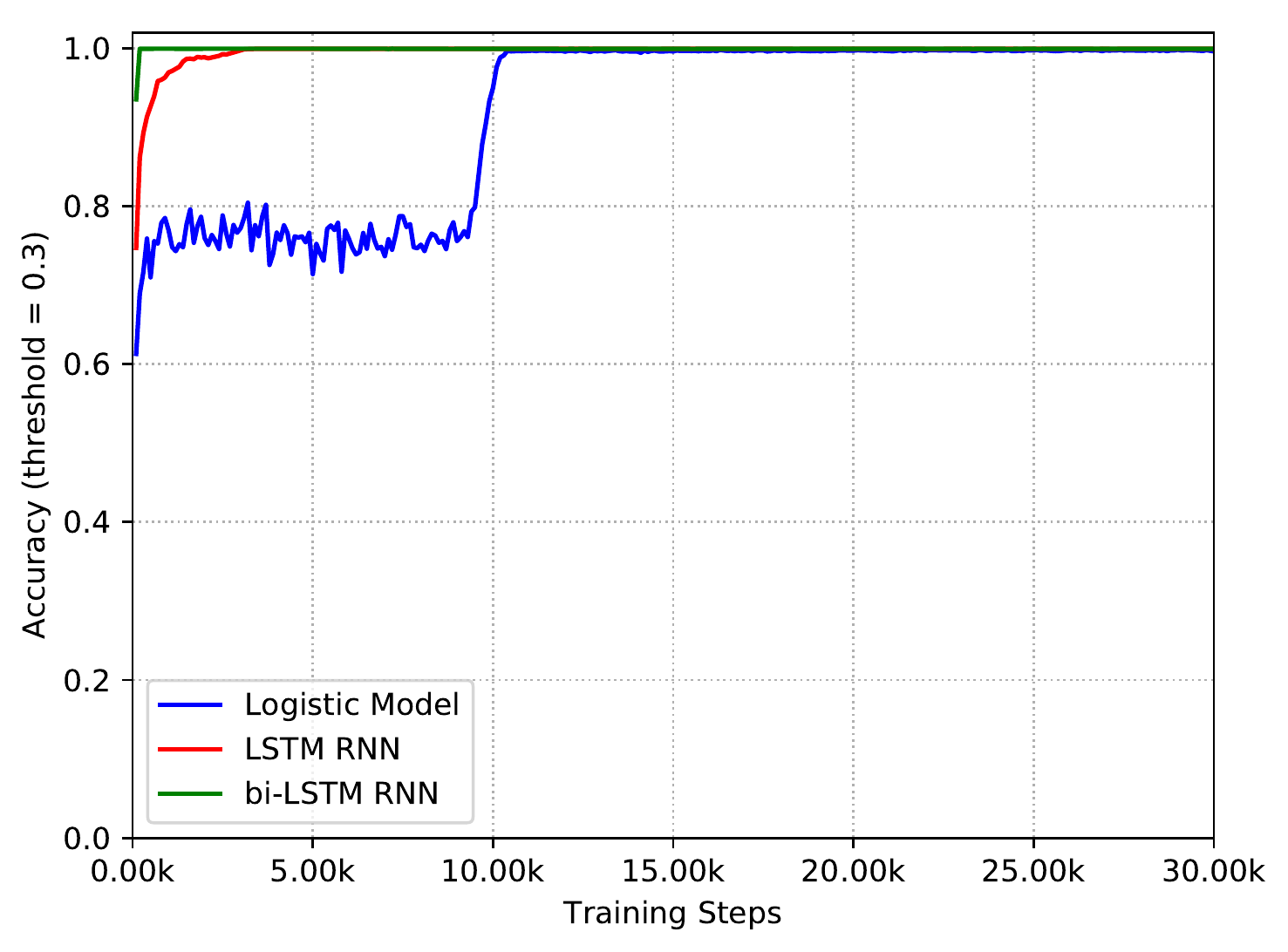}}\subfloat[Precision\label{fig:Precision_validation}]{
\centering{}\includegraphics[scale=0.53]{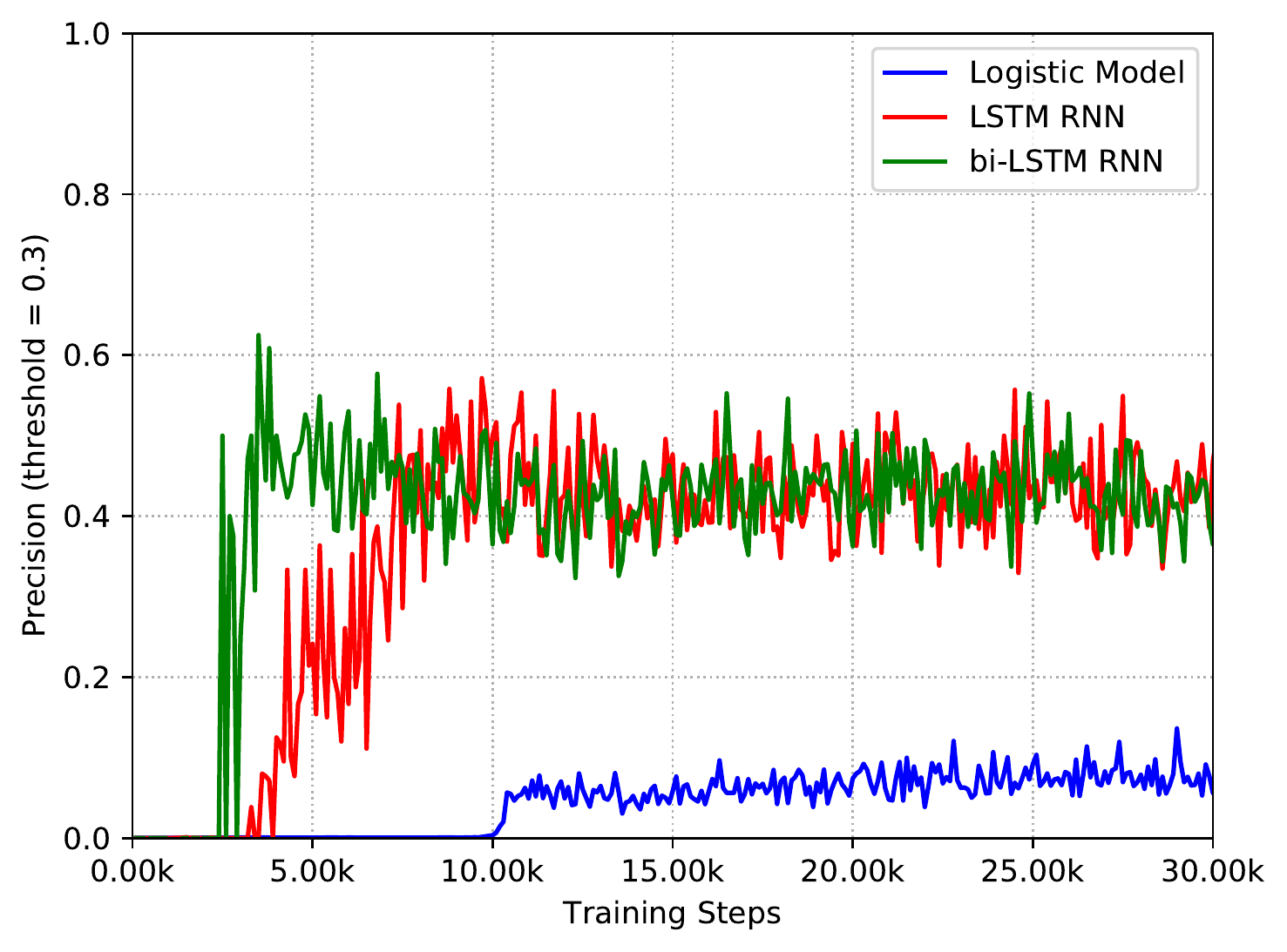}}
\par\end{centering}
\begin{centering}
\subfloat[Recall\label{fig:recall_validation}]{
\centering{}\includegraphics[scale=0.53]{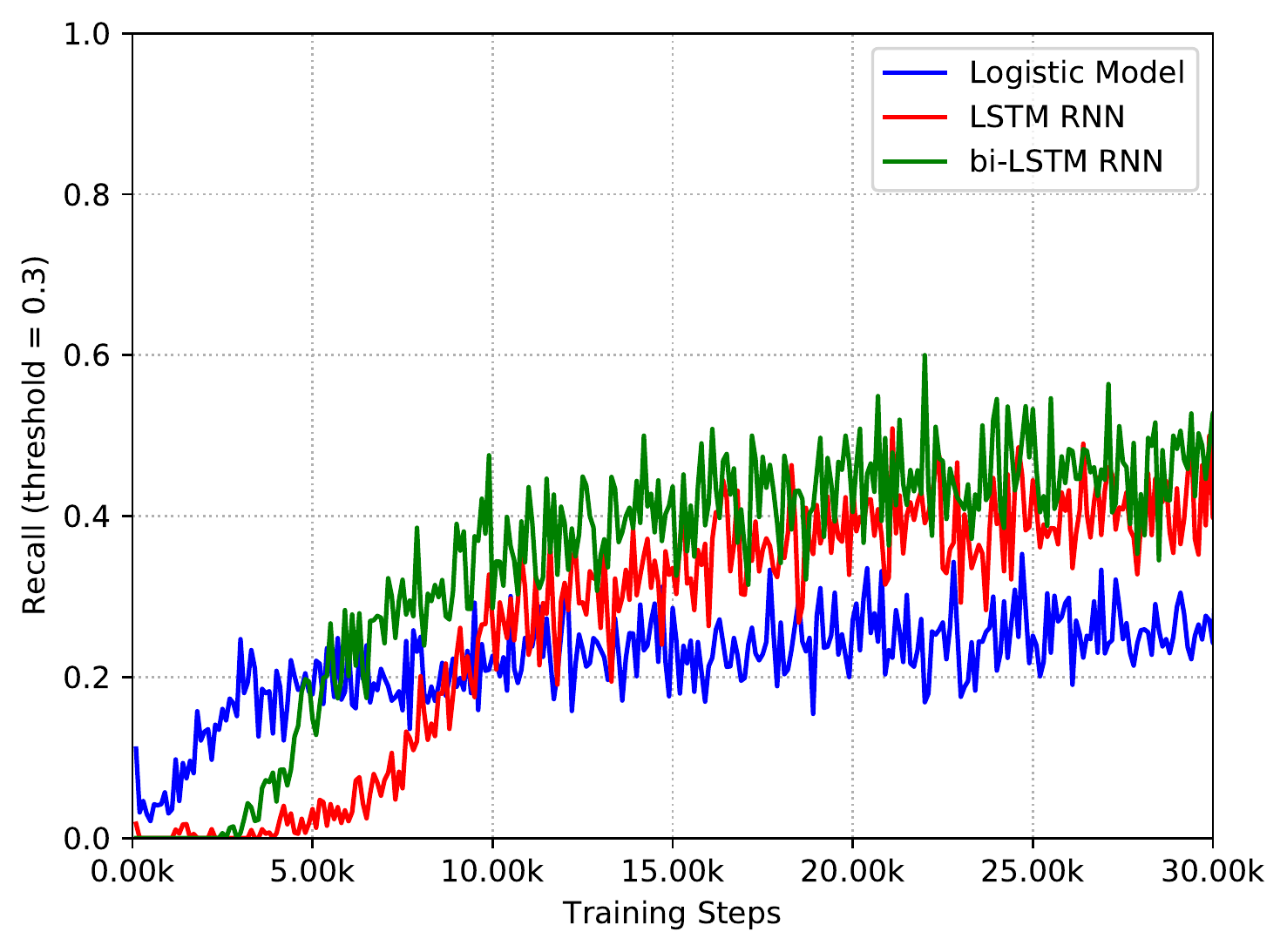}}\subfloat[AUC\label{fig:AUC_validation}]{
\centering{}\includegraphics[scale=0.53]{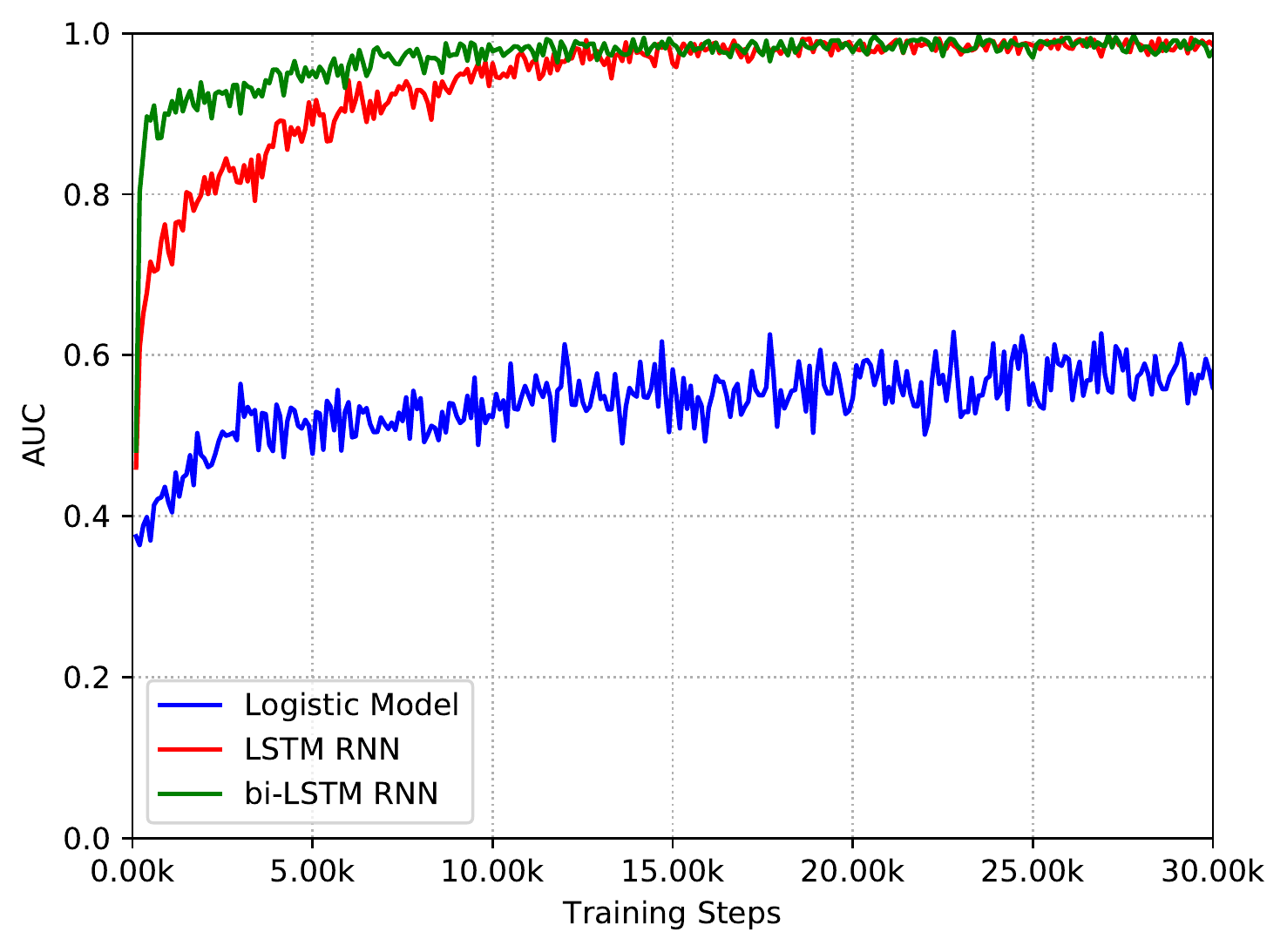}}
\par\end{centering}
\begin{spacing}{0.9}
\centering{}\begin{threeparttable}\begin{tablenotes} \item \textit{\scriptsize{}Notes:}{\scriptsize{}
The figures benchmark the accuracy, precision, recall and AUC (area
under the curve) statistics of the bi-directional RNN, against a unidirectional
LSTM RNN and a logistic model.}\end{tablenotes} \end{threeparttable} 
\end{spacing}
\end{figure}

Figures \ref{fig:Performance-of-BidirectionalRNNBenchmarks} compares
the accuracy, precision, recall and AUC (area under the curve) statistics
of the model against two benchmark specifications.\footnote{The statistics are computed on a validation dataset that is held-out
separately from the training dataset.} The first is a unidirectional LSTM RNN, which is exactly the same
as the preferred model but without the forward recurrence. The second
is a flexible logistic model, which specifies the probability of a
user $i$ purchasing brand $b$ on day $t$ as a semi-parametric logistic
function of the ad-impressions and price-indexes on the same day (i.e.,
$\Pr\left(Y_{ibt}=1\right)=\textrm{logistic}\left(\mathbf{x}_{it},\mathbf{p}_{t};\Omega\right)$).
Looking at the results, we see the bi-directional RNN has the highest
AUC amongst the models; and has accuracy, precision and recall statistics
that are comparable or higher. The poor performance of the logistic
model in particular emphasizes the importance of accounting for dependence
over time to fit the data. The plots also show the speed of convergence
of the models as a function of the number of training steps; the bi-directional
RNN converges faster in fewer training steps. This is helpful in production,
which typically requires frequent model updating.\footnote{To get a sense for this, the training times for 30,000 steps for the
3 models on our cluster are 11.21 hrs (bi-directional RNN); 9.68 hrs
(unidirectional RNN); 12.48 hrs (logistic) hrs respectively. }

Tables \ref{tab:Prediction-Error} also benchmarks the algorithm presented
in \ref{alg:shap_dist} for distributed computation of Shapley values.
Recall this algorithm shifts from exact computation of Shapley values
to a Monte Carlo simulation approximation when the number of ad-positions
over which to allocate credit is large. This ``mixed'' method improves
computational speed, which is important for high-frequency reporting
of results in deployment. To assess the performance of this algorithm,
we pick 6,000 orders from $t=T$ and run the algorithm on these data
for various configurations. The experiment is repeated for each configuration
5 times, and the average across the 5 reps is reported.\footnote{The computational environment uses a \texttt{Spark} cluster with \texttt{Spark}
2.3 by \texttt{pyspark}, running \texttt{TensorFlow} v1.6, with an
8 core CPU, 100 workers and 8 GB memory per worker without the GPU.} The first row in the table reports on the number of orders we are
able to attribute per minute using the three methods: the mixed method
is about 2,300\% faster than exact computation, and about 14\% faster
than a simulation-only method. The second row documents this efficiency
gain does not come at the cost of high error: the average error in
the mixed method is low relative to exact computation, and a order
of magnitude smaller than using full simulation.\footnote{We compute error as a mean squared difference over the $l=1,..6000$
orders in the total attributed value (\texttt{line $20$} in algorithm
\ref{alg:shap_dist}) for the evaluated algorithm $\left(SA_{l}\right)$
relative to that from the exact algorithm $\left(SA_{l}^{*}\right)$,
i.e. $\textrm{err}=\sqrt{\tfrac{1}{6000}\sum_{l=1}^{6000}\left(SA_{l}-SA_{l}^{*}\right)^{2}}$.}
\begin{table}
\begin{centering}
\caption{Processing Efficiency and Prediction Error when using Exact, Approximate
and Mixed Shapley Value Computation\label{tab:Prediction-Error}}
\par\end{centering}
\medskip{}

\centering{}%
\begin{tabular}{|r|c|c|c|}
\hline 
\textbf{\footnotesize{}Algorithm} & \textbf{\footnotesize{}Exact} & \textbf{\footnotesize{}Approximate} & \textbf{\footnotesize{}Mixed}\tabularnewline
\hline 
{\footnotesize{}Orders processed per minute} & {\footnotesize{}4.2} & {\footnotesize{}88.85} & {\footnotesize{}101.24}\tabularnewline
\hline 
{\footnotesize{}Error} & $-$ & {\footnotesize{}0.3190} & {\footnotesize{}0.0064}\tabularnewline
\hline 
\end{tabular}
\end{table}

\paragraph*{Model Results}

\begin{figure}
\begin{centering}
\centering \caption{Distribution Across Orders of How Much of the Probability of Purchase
is Incrementally Driven by Advertising\label{fig:Incremental-Orders-Due-toAdvertising}}
\par\end{centering}
\begin{centering}
\includegraphics[scale=0.6]{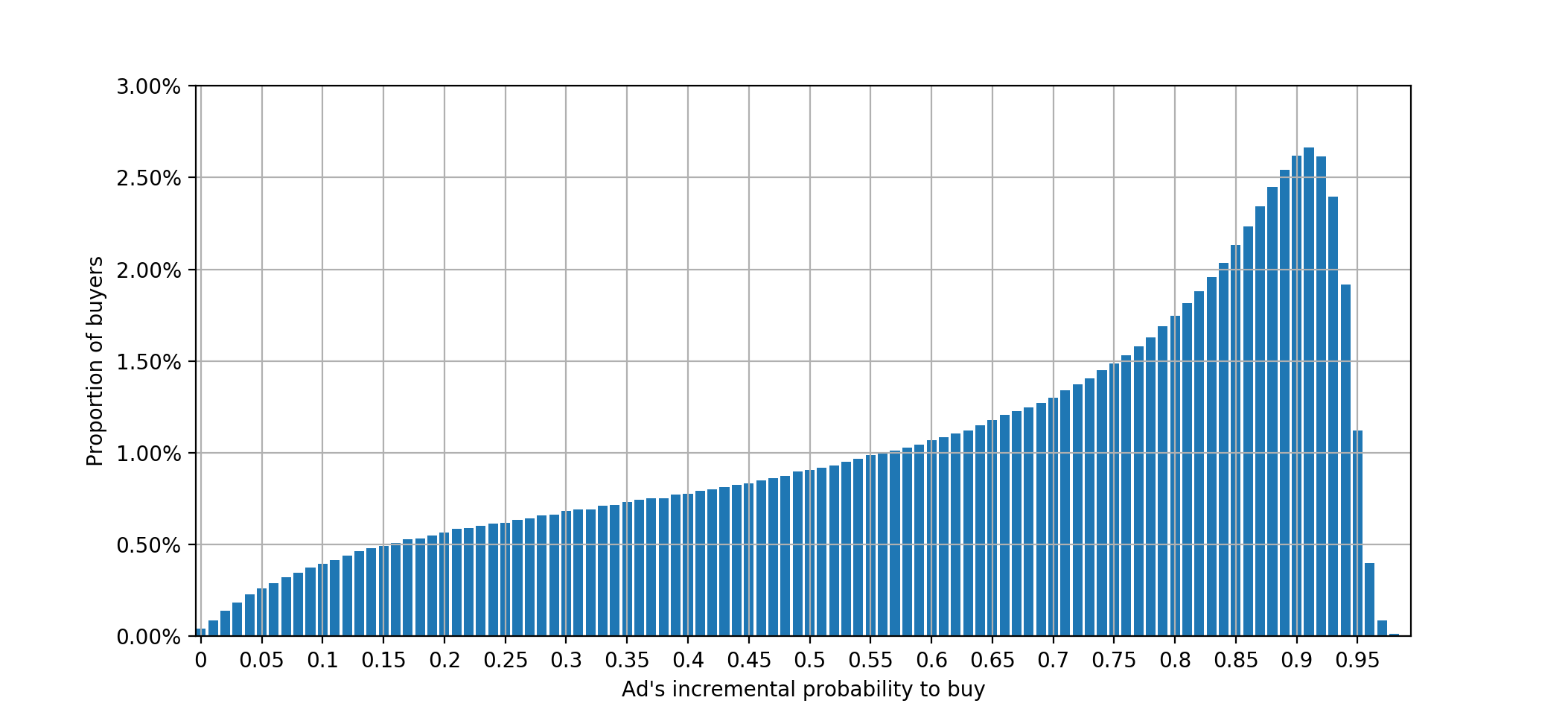}
\par\end{centering}
\begin{spacing}{0.9}
\centering{}\begin{threeparttable}\begin{tablenotes} \item \textit{\footnotesize{}Notes:}{\scriptsize{}
The Figure shows how much of the probability of purchase is driven
by advertising. For each order on day $T=15$, we compute the predicted
probability of that order occurring with and without advertising exposure,
and take the difference as the incremental probability associated
with advertising. We then compute the ratio of incremental to the
total predicted probability with advertising. The figure presents
a histogram of this ratio across all orders.}\end{tablenotes} \end{threeparttable} 
\end{spacing}
\end{figure}

To explore the results from the model, we first discuss Figure \ref{fig:Incremental-Orders-Due-toAdvertising}
which emphasizes the importance of accounting for the incrementality
of advertising in the allocation of orders. The figure assesses how
much of the probability of an observed order occurring is driven by
advertising. For each observed order in the cell-phone category on
day $T=15$, we compute the predicted probability of that order occurring
with and without advertising exposure, and take the difference as
the incremental probability associated with advertising. We then compute
the ratio of incremental to the total predicted probability with advertising.
The figure presents a histogram of this ratio across all orders. There
is a substantial spread from $0-1$, with some orders that seems driven
primarily by advertising, and some which would have occurred anyway
irrespective of the brand's advertising. It is this counterfactual
outcome of what would the order would have been in the absence of
advertising that the incrementality based allocation seeks to reflect.
\begin{figure}
\begin{centering}
\centering \caption{Empirical CDF of Computed Shapley Values\label{fig:Empirical-CDF-ofShapleyValues}}
\par\end{centering}
\begin{centering}
\includegraphics[scale=0.65]{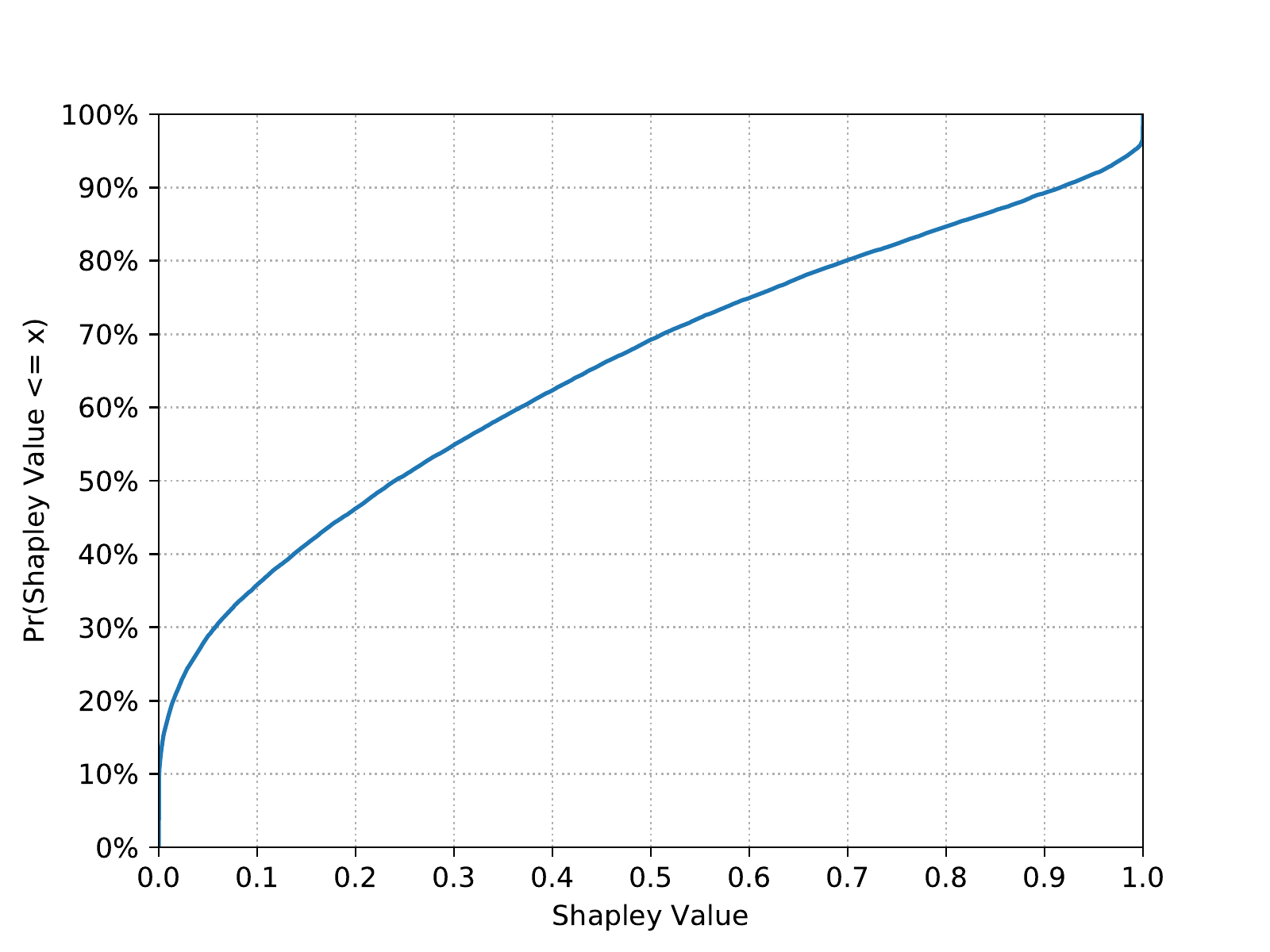}
\par\end{centering}
\begin{spacing}{0.9}
\centering{}\begin{threeparttable}\begin{tablenotes} \item \textit{\scriptsize{}Notes:}{\scriptsize{}
The Figure shows the eCDF of the Shapley values computed on the basis
of the RNN model for all the orders on day-15. The median Shapley
value is about 0.25, suggesting the median ad-position to which exposure
occurs contributes about 25\% to the incremental benefit from advertising.
About 70\% of the Shapley values are above 0.5, suggesting these ad-positions
contribute more than half of the the incremental benefit generated
from advertising.} {\scriptsize{}About 15\% of the Shapley values
are above 0.8, suggesting these ad-positions contribute more than
80\% of the the incremental benefit generated from advertising.}\end{tablenotes}
\end{threeparttable} 
\end{spacing}
\end{figure}
\begin{figure}
\begin{centering}
\centering \caption{Shapley Values from RNN Model at each Ad-position, Averaged Across
all Orders in which that Ad-position was the ``Last-clicked''\label{fig:Shapley-Values-vsLastClicked}}
\par\end{centering}
\begin{centering}
\includegraphics[scale=0.68]{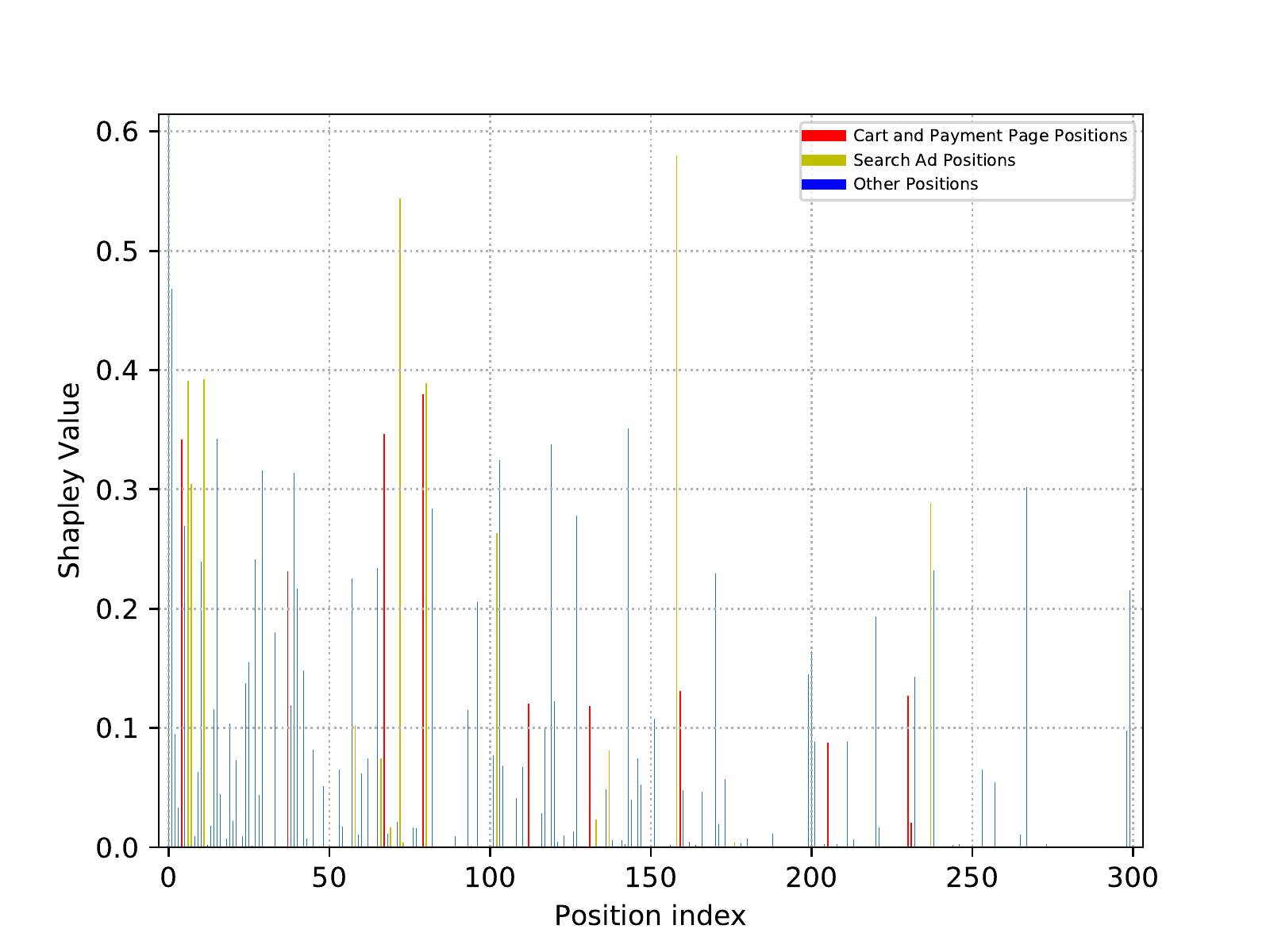}
\par\end{centering}
\begin{spacing}{0.9}
\centering{}\begin{threeparttable}\begin{tablenotes} \item \textit{\scriptsize{}Notes:}{\scriptsize{}
The Figure shows the Shapley Value from the RNN model at each ad-position
indexed on the $x$-axis, average across all orders on day-15 for
which that position was the last clicked. This allows benchmarking
the Shapley Value based attribution against ``last-click'' attribution,
which allocates 100\% of the credit for the order to the last-clicked
ad-position. The Shapley values are all seen to be <1, showing that
under the model, the last-clicked ad-positions do not obtain full
credit. To the extent that the Shapley values are all less than 0.6,
the RNN model suggests that last-clicked ads contribute upto a maximum
of 60\% to the incremental conversion generated by advertising. Finally,
cart and payment page positions, which may get a lot of credit under
``last-click'' or ``last-visit'' attribution schemes on eCommerce
sites, are seen to not be allocated a lot of credit by the model.}\end{tablenotes}
\end{threeparttable} 
\end{spacing}
\end{figure}

Figure \ref{fig:Empirical-CDF-ofShapleyValues} shows the empirical
CDF of the Shapley values computed on the basis of the RNN model for
all the orders on day $T=15$. The median Shapley value is about 0.25,
suggesting the median ad-position to which exposure occurs contributes
about 25\% to the incremental benefit from advertising. About 70\%
of the Shapley values are above 0.5, suggesting these ad-positions
contribute more than half of the the incremental benefit generated
from advertising. About 15\% of the Shapley values are above 0.8,
suggesting these ad-positions contribute more than 80\% of the the
incremental benefit generated from advertising. This shows the Shapley
values have discriminatory power $-$ it helps identifying top ad-positions
that contribute most to observed outcomes.

Finally, Figure \ref{fig:Shapley-Values-vsLastClicked} compares the
credit allocation based on Shapley Values to rule-based ``Last-clicked''
attribution. To do this, Figure \ref{fig:Shapley-Values-vsLastClicked}
shows the Shapley Value from the RNN model at each ad-position indexed
on the $x$-axis, averaged across all orders on day $T=15$ for which
that position was the last clicked. This allows benchmarking the Shapley
Value based attribution against ``last-click'' attribution, which
allocates 100\% of the credit for the order to the last-clicked ad-position.
The Shapley values are all seen to be <1, showing that under the model,
the last-clicked ad-positions do not obtain full credit. To the extent
that the Shapley values are all less than 0.6, the RNN model suggests
that last-clicked ads contribute upto a maximum of 60\% to the incremental
conversion generated by advertising. Further, cart and payment page
positions (which correspond to ads shown on these positions), which
may get a lot of credit under ``last-click'' or ``last-visit''
attribution schemes on eCommerce sites, are seen to not be allocated
a lot of credit by the model.

\begin{sidewaysfigure}
\begin{centering}
\centering \subfloat[Impression Share by Ad-Position\label{fig:ImpressionShareAd-Position}]{
\begin{centering}
\includegraphics[scale=0.65]{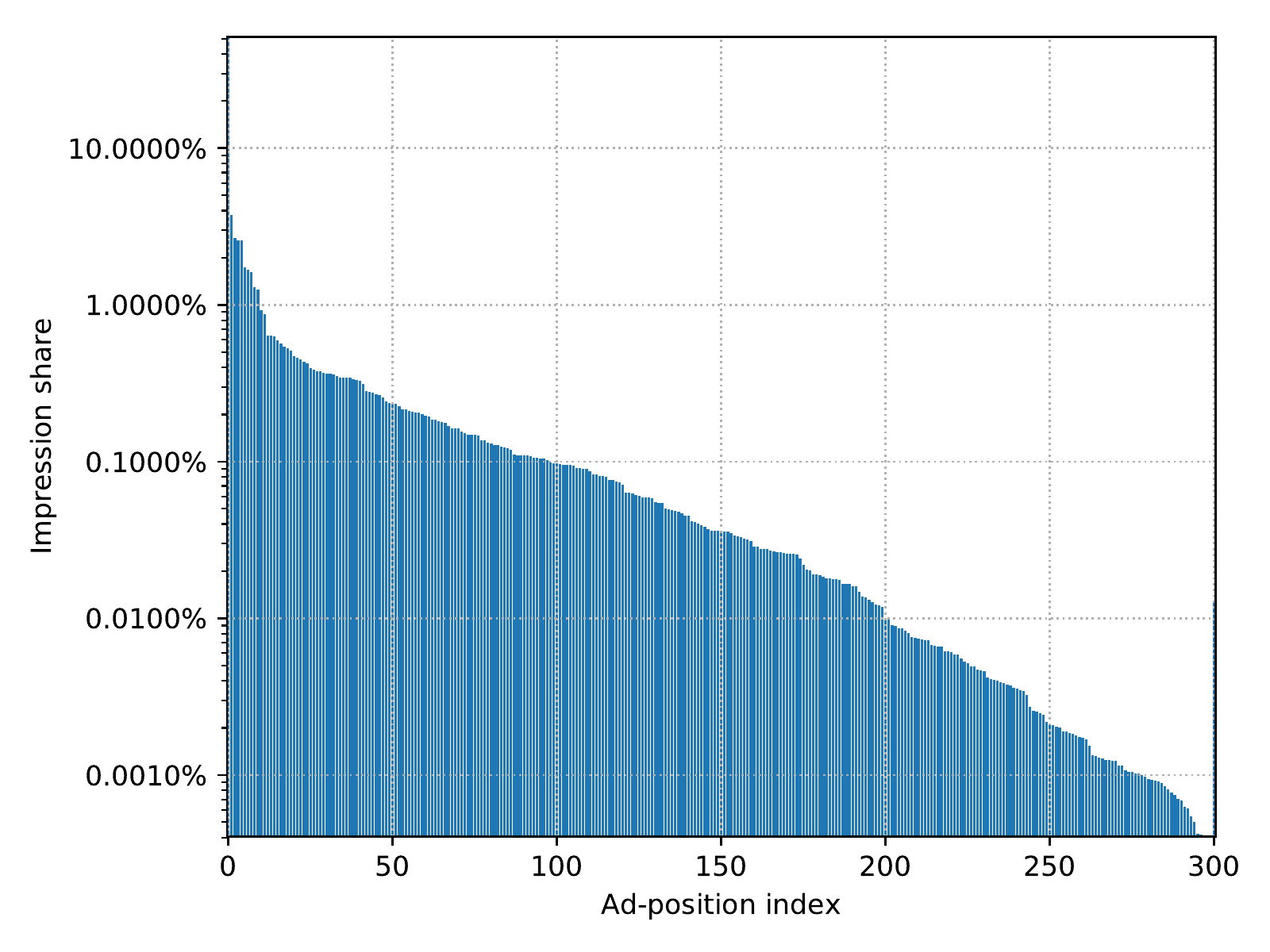}
\par\end{centering}
}\subfloat[Shapley Values by Ad-Position\label{fig:ShapleyValuesAd-Position}]{
\centering{}\includegraphics[scale=0.7]{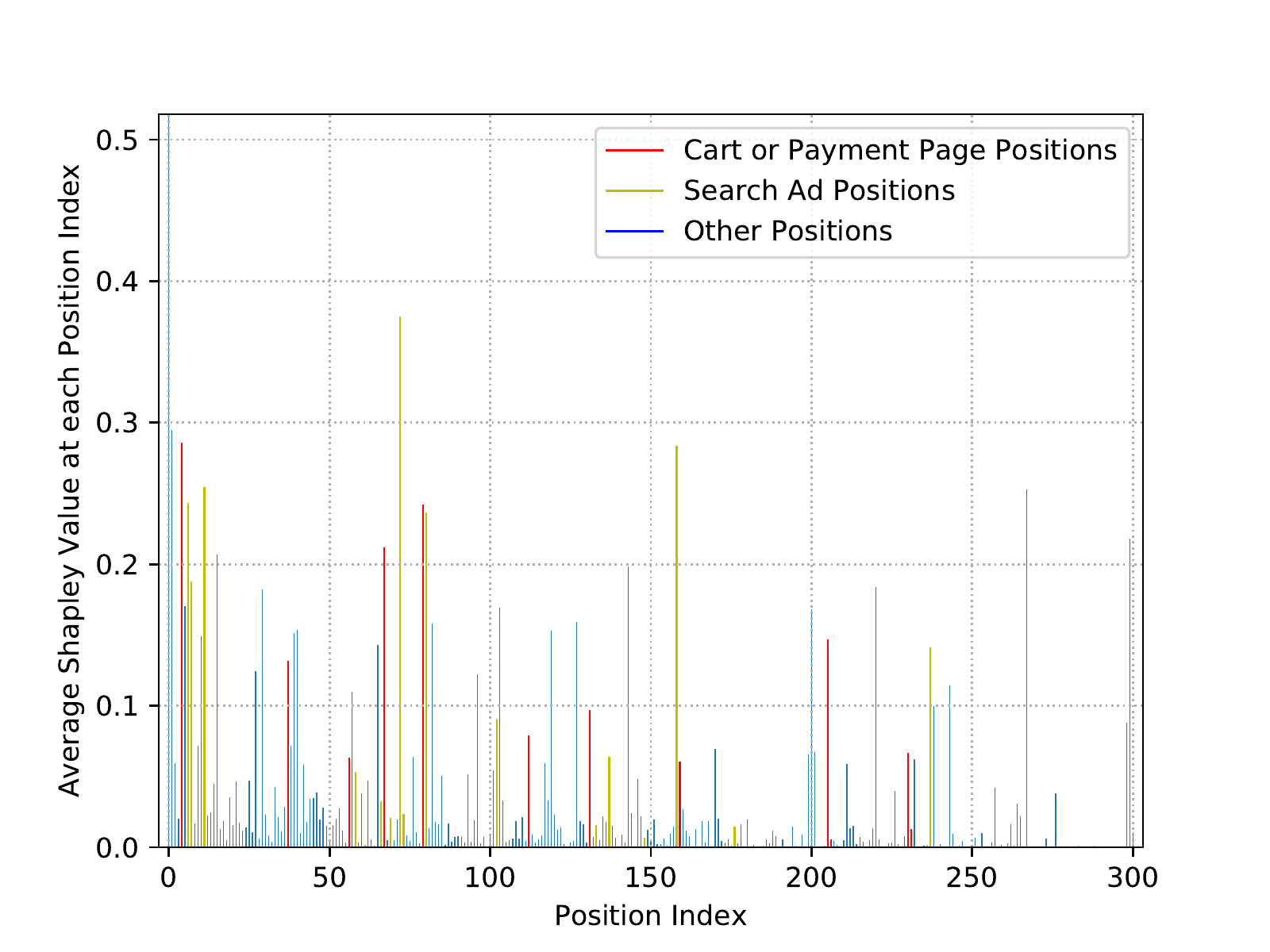}}
\par\end{centering}
\begin{spacing}{0.9}
\centering{}\begin{threeparttable}\begin{tablenotes} \item \textit{\scriptsize{}Notes:}{\scriptsize{}
The left panel of the figure shows the proportion of ad-impressions
in the training data across the ad-positions. The ad-positions are
indexed from $1-301$ in order of their share of total impressions.
The right panel of the figure shown the average (across orders) of
the Shapley Values for the same ad-positions. Comparing the two figures,
we can see that the Shapley Values do not simply map out the intensity
of impressions.} \end{tablenotes} \end{threeparttable} 
\end{spacing}
\end{sidewaysfigure}

As a final assessment, we compare Figure \ref{fig:ImpressionShareAd-Position}
which shows the proportion of ad-impressions in the training data
across the ad-positions to Figure \ref{fig:ShapleyValuesAd-Position}
which shows the average (across orders) of the Shapley Values for
the same ad-positions. The ad-positions are indexed from $1-301$
in order of their share of total impressions, so impressions of ad-position
$\#k$ is higher than impressions of ad-position $\#k+1$, and so
on in both figures. Comparing the two figures, we can see that the
distribution of Shapley Values across positions does not follow the
same pattern as that of impressions, suggesting that the effect picked
up by the model is not purely driven by the intensity of advertising
expenditures by advertisers (which drive impressions). Further, we
observe that some positions that receive fewer ad-impressions have
higher Shapley Values than those that receive higher impressions.
This suggests that advertising expenditure allocations overall may
not be optimal from the advertiser\textquoteright s perspective, and
could be improved by better incorporation of attribution using a model
such as the one presented here. A more formal assessment of this issue
however requires a method for advertiser budget allocation across
ad-positions, which is outside of the scope of this paper.

\section{Implementation and Extension to Larger Scale}

For actual implementation, the model has to be scaled to all brands
and all categories on \texttt{JD.com}. From a model training and updating
perspective, it is intractable to maintain a separate model for each
product category (more than 175). For production, we extend the model
presented above to accommodate all product categories in one unified
framework. This model has larger scale, so we impose some parameter
restrictions to reduce the dimensionality of the problem. First, we
allow categories to have separate parameters, but restrict the weights
for all brands \textit{within a} product category to be similar. Second,
on the basis of pre-training data, we create a set of features that
characterize each brand (e.g., brand rank within\texttt{ JD.com}),
and include them as covariates that shift the intercept of the output
layer. This allows for heterogeneity across brands within each product
category. Third, for each user $i$, brand $b$, and day $t$, we
include in the input vector $\mathbf{x}_{it}$ the ad-impressions
of brand $b$ at all the $K$ ad-positions as before; but summarize
the competitive ad-impressions by including only (a) the ad-impressions
of all brands other than $b$ in the same product category across
the $K$ positions; and (b) the ad-impressions of all brands in other
categories across the $K$ positions. Thus, $\mathbf{x}_{it}$ in
this model is a $3K\times1$ dimensional vector with the first $K$
entries corresponding to impressions of the focal brand; the next
$K$ corresponding to all other brands in the same product category
as the focal brand; and the last $K$ corresponding to all brands
in other categories. Fourth, in order to increase the informativeness
of user impression of ads, in some specifications, we include information
on whether the user clicked on ads into the response model. Finally,
to address the issue of selection more directly, we use the pre-training
dataset to develop a predicted baseline propensity of each user $i$
to buy a particular brand $b$, using as features his characteristics
$d_{i}$. We include the predicted baseline propensities into the
unified model as controls. Due to business confidentiality reasons,
we do not reveal the exact details of this implementation.

\section{Conclusions\label{sec:Conclusion}}

A practical system for data-driven MTA for use by ad publishing platform
is presented. The system combines a flexible response model trained
on user-level data with Shapley values for ad-types for attribution.
A bi-directional RNN customized to modeling user purchase behavior
and ad-response, is developed as the response model, which has the
advantage of being semi-parametric, reflective of several salient
aspects of ad-response, and able to handle high dimensionality and
long-term dependence. The use of the Shapley value provides a way
to allocate credit at a disaggregate level in a way that respects
the sequential nature of advertising response. The use of the Shapley
value is based on fairness considerations taking the advertising policies
by the advertisers as given. It is possible that advertisers re-optimize
their advertising policies in response to the allocations. The optimal
allocation contract that endogenizes the equilibrium response of advertisers
remains an open question (see, for instance, \citet{Abhisheketal2017,Berman2018}).

\newpage{}

\baselineskip=.8pc \parskip=0in \bibliographystyle{ecta}

\end{document}